\pdfoutput=1
\documentclass[journal]{IEEEtran}
\usepackage{graphicx}
\usepackage[justification=centering]{caption}
\usepackage{verbatim}
\usepackage{cite}
\usepackage{multirow}
\usepackage{colortbl,booktabs}
\usepackage{epstopdf}
\usepackage{stfloats}
\usepackage{cite}
\ifCLASSINFOpdf
\else
\fi
\usepackage{amsmath}

\usepackage{amssymb}

\begin{document}
%

\title{Tensorizing GAN with High-Order Pooling for Alzheimer's Disease Assessment}

\author{Wen~Yu$^{\dagger}$,
        Baiying~Lei$^{\dagger}$,
        Michael~K.~Ng,
        Albert~C.~Cheung,
        Yanyan~Shen,
        and~Shuqiang~Wang
        \thanks{  $^{\dagger}$The first two authors contributed equally to this work.}
\thanks{ W. Yu, Y. Shen and SQ. Wang are with Shenzhen Institutes of Advanced Technology, Chinese Academy of Sciences, Shenzhen 518060, China. E-mail: sq.wang@siat.ac.cn.}
\thanks{B. Lei is with School of Biomedical Engineering, Shenzhen University, Shenzhen 518060, China. E-mail:leiby@szu.edu.cn}
\thanks{Michael K. Ng is with Department of Mathematics,
	The University of Hong Kong, Pokfulam, Hong Kong. E-mail: mng@maths.hku.hk}
\thanks{Albert C. Cheung is with Hong Kong University of Science and Technology, Hong Kong, E-mail: albertccheung@yahoo.com}}


\markboth{  }%
{Shell \MakeLowercase{\textit{et al.}}: Bare Demo of IEEEtran.cls for IEEE Journals}

\maketitle

\begin{abstract}
It is of great significance to apply deep learning for the early diagnosis of Alzheimer's Disease (AD). In this work, a novel tensorizing GAN with high-order pooling is proposed to assess Mild Cognitive Impairment (MCI) and AD. By tensorizing a three-player cooperative game based framework, the proposed model can benefit from the structural information of the brain.
By incorporating the high-order pooling scheme into the classifier, the proposed model
can make full use of the second-order statistics of the holistic Magnetic Resonance Imaging (MRI) images.
To the best of our knowledge, the proposed
{\tt T}ensor-train, {\tt H}igh-pooling and {\tt S}emi-supervised learning based GAN (THS-GAN)
is the first work to deal with classification on MRI images for AD diagnosis. Extensive experimental
results on Alzheimer's Disease Neuroimaging Initiative (ADNI) dataset are reported to demonstrate that the proposed THS-GAN
achieves superior performance compared with existing methods,
and to show that both tensor-train and high-order pooling can enhance classification performance.
The visualization of generated samples also shows that the proposed model can generate plausible samples for semi-supervised
learning purpose.
\end{abstract}


\IEEEpeerreviewmaketitle

\section{Introduction}\label{introduction}

%
\IEEEPARstart{A}{lzheimer's Disease (AD)}  is an irreversible and chronic neurodegenerative disease with progressive impairment of memory and other mental functions. It is estimated to be the third leading cause of death, after heart disease and cancer \cite{ADimportant}.   According to the World Alzheimer Report \cite{adReport}, the total estimated prevalence of AD was around 50 million worldwide in 2018, and the number will  increase to 152 million by 2050.  AD is caused by abnormal deposits of protein in the brain that destroys cells in the regions that control memory and mental functions. To date, AD is incurable but preventable. Early diagnosis of AD is crucial for timely therapy to slow the progression of the disease.  Currently, the clinical diagnosis of AD heavily depends on clinical history\cite{AD_significant2}. The diagnosis procedure is time-consuming and requires extensive clinical training and experience for neurologists. Therefore,   accurate AD assessment in its earliest stage by utilizing deep learning is highly desirable.

T1-MRI image is an important biomarker for AD diagnosis in routine clinical practice. Early work for AD diagnosis using MRI images primarily focused on traditional machine learning techniques \cite{AD_old12}\cite{AD_old16}, which heavily relied on specific assumptions about brain structural abnormalities, such as regional cortical thickness, hippocampal volume, and gray matter volume.  The performance of these manual feature extraction methods is limited since they require advanced clinical domain knowledge and complicated preprocessing steps. Therefore, they tend to be time-consuming and subjective.  Besides, the brain is a huge network with complicated connections. The disease-related structure changes are subtle and scattered throughout the entire brain in different tissues.  These kinds of patterns are difficult to learn since not all morphological abnormalities related to AD can be captured accurately, and the extracted Regions Of Interest (ROI) or voxel features are processed independently. Hence these features are unable to express the internal brain connections sufficiently.

Recent advances in deep learning \cite{DeepLearning} \cite{3D_GAN_PET} have explosive popularity in computer vision and various medical applications.  Instead of manually extracting features according to domain-specific knowledge, deep learning can discover the discriminant representations of images by incorporating the feature extraction into the task learning process.  However, most existing methods can only utilize the labeled data in a supervised manner.   Annotation of MRI images is laborious and costly, which requires clinical confirmation with great effort by experts. As a result, only small amounts of labeled MRI images are available for AD assessment, and the unlabeled MRI images can not be used directly.

Generative Adversarial Network (GAN) has attracted much attention as it is capable of generating data without explicitly modeling the probability density function. It is intelligent for the discriminator to incorporate unlabeled data into the training process by utilizing the adversarial loss \cite{GanReview}. Furthermore, GAN has been proven to be feasible in data augmentation, image-to-image translation, and Semi-Supervised Learning (SSL).
To make full use of both labeled and unlabeled MRI images, Semi-Supervised GAN (SS-GAN) \cite{gan_goodfellow,semiGAN,gan_hard_train_57,SSgan}
can be adopted.
In this paper, our primary goal is to leverage GAN to characterize the high-order distribution of MRI images for semi-supervised classification. In particular, we discovered that the recently introduced triple-GAN could alleviate the instability and incompatible problems
of the SS-GAN \cite{SSgan}. Triple-GAN designed a three-player cooperative game instead of the conventional two-player competition game by introducing the auxiliary classifier network based on generator and discriminator.  Inspired by this,
our model exploits the three-player cooperative game for modeling MRI images to assess  MCI and AD.

Based on these observations, in this paper, we propose a novel Tensorizing GAN with High-order pooling to assess MCI and AD.
More specifically, in order to stabilize the training of GAN and speed up the convergence, the proposed model
utilizes the compatible learning objects of the three-player cooperative game.
Our proposed model is called THS-GAN, i.e., {\tt T}ensor-train decomposition, {\tt H}igher-order pooling,
and {\tt S}emi-supervised learning are employed in the proposed GAN model.
Instead of vectorizing each layer as conventional GAN, the tensor-train decomposition is applied to all layers in classifier and discriminator, including fully-connected layers and convolutional layers. Thus the number of parameters
can be reduced significantly.
Besides, in such a tensor-train format, our model can benefit from the structural information of the brain.
Moreover, compared with the first-order pooling, the high-order pooling module can extract more significant features by making full use of the second-order statistics of the holistic MRI image. Thus our model also exploits Global Second-order Pooling (GSP) block
as a high-order pooling module
in the classifier.   In particular,  GSP block can capture the long-range dependencies of features at distant positions by computing all pairwise channel correlations of the 4D feature-maps extracted by 3D-DenseNet. Thus both GSP and 3D-DenseNet are integrated into the classifier to enhance salient feature channels and suppress less-useful feature channels. As a result, useful features related to anatomical abnormalities
are extracted in a self-attention manner to improve the performance of classification. The contributions of this paper are summarized as follows:
\begin{enumerate}
	\item      By tensorizing the three-player cooperative game based framework,
	the proposed model can benefit from the structural information
	of the brain.
	\item      The proposed  THS-GAN leverages the high-order pooling to make full use of the second-order statistics of the holistic MRI images. The long-range dependencies between slices of different directions can be captured effectively. Thus more significant features can be extracted automatically in a self-attention manner to boost the predictive performance.
	\item    The THS-GAN model is designed to assess MCI and AD in a semi-supervised manner to take advantage of both labeled and unlabeled MRI images.
\end{enumerate}

The rest of this paper is organized as follows. We review
the related work in Section \ref{Related_work}.
In Section \ref{method}, we present the proposed THS-GAN in detail.
In Section \ref{experimental_results},  THS-GAN is  tested with various configurations and experimental results are presented to demonstrate its advantage.
Finally, concluding remarks and future work are discussed in
Section \ref{Conclusions}.

\begin{figure*} 
	\includegraphics[width=\linewidth]{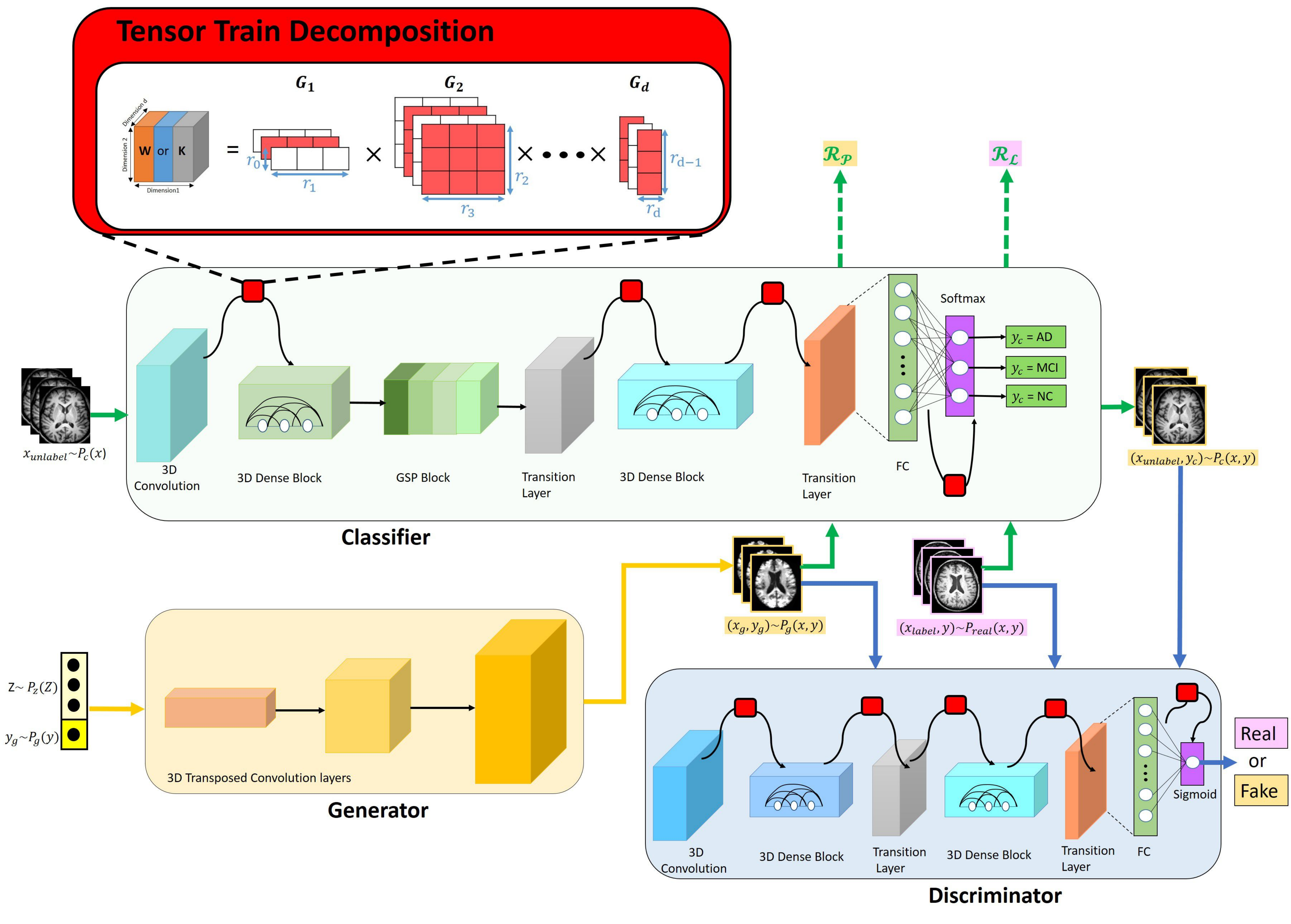}
	\caption{An illustration of THS-GAN (best view in color). Real and Fake are the adversarial losses. $\mathcal{R}_{\mathcal{L}}$  and $\mathcal{R}_{\mathcal{P}}$ denote the cross-entropy loss for supervised learning for real data and generated data respectively.  $\mathcal{R}_{\mathcal{L}}$  and $\mathcal{R}_{\mathcal{P}}$ are unbiased regularizations that ensure the consistency between $p_{g}$,$p_{c}$ and $p_{\text{real}}$, which are the distributions defined by the generator, classifier and true data respectively.}
	\label{fig_triple_gan}
\end{figure*}
\section{Related work} \label{Related_work}
The current AD diagnosis model can be categorized into two types: the traditional machine learning-based approach and the deep learning-based approach.

The traditional machine learning techniques can be divided further into three categories: Voxel-based approach,  ROI-based approach, and patch-based approach.  Although the voxel-based approach \cite{voxel_AD1} is intuitive and straightforward in terms of interpretation, the process of classification is computationally expensive since the voxel-wise features are of extremely high dimensionality, and the classification performance will deteriorate due to the ``curse of dimensionality'' \cite{old16}.    For ROI-based approach \cite{AD_old12}, the ROIs are segmented by prior hypothesis, but the abnormal regions related to AD may not fit the predefined ROIs ideally in practice,  and the features extracted from ROIs are very coarse in the sense that they can't sufficiently represent all subtle changes involved in the brain diseases. As a result, the representation power of ROI features is limited.    Patch-based approach dissected brain areas into small 3D-patches, followed by extracting features from each selected patch individually, and then the features are combined hierarchically in a classifier level \cite{patch_AD2}. However, the features extracted by these methods neglect the correlated variations of the whole brain structure affected by AD in other regions. Besides, the extraction of these handcrafted features heavily depend on how well the images are registered and segmented, which often require the domain expert knowledge.

In the application domain of AD diagnosis, the previous deep learning studies focused on two directions: (1) CNN is utilized for supervised classification \cite{ad3Dcnn}, primarily by using large-scale annotated data sets. (2) Unsupervised GAN is exploited for data synthesis or image-to-image translation \cite{AD_GAN_MRI_PET} \cite{ganTranslateMRI}.  In the first approach, Islam et al. \cite{AD_DL_28} presented a method based on 2D-DenseNet. The MRI images is sliced in three directions (axial, coronal, and sagittal). Then three parallel 2D-DenseNets are evaluated on MRI slices separately. Finally, the results are fused for AD diagnosis.  However, the way of converting a 3D-image into a series of 2D-slices causes CNNs to disregard the spatial information of 3D space, and different slicing methods lead to loss of   features. Thus many studies focus on 3D-CNN instead of 2D to alleviate this issue. For instance, Payan et al. \cite{ad3Dcnn} utilized 3D-convolutions \cite{compare8} combined with a sparse auto-encoder, which yielded better performance than 2D-convolutions on slices. In the second approach, Pan et al. \cite{AD_GAN_MRI_PET} imputed the missing PET images by learning bi-directional mappings between MRI and PET via 3D-cGAN. Then, based on the complete MRI and PET (after imputation), they develop a landmark-based multi-modal multi-instance learning method (LM3IL) for AD diagnosis.  Karim et al. \cite{ganTranslateMRI} proposed the Cycle-MedGAN framework based on the traditional Cycle-GAN with new non-adversarial losses for PET to CT translation. Wang et al. \cite{3D_GAN_PET} proposed a 3D auto-context-based locality adaptive multi-modality generative adversarial network model (LA-GANs) to synthesize the high-quality FDG PET image from the low-dose one with the MRI images that provide anatomical information.

In this paper, our approach is different from the previous GAN applications on AD diagnosis that focus on image synthesis and image-to-image translation. Our aim is to enhance GAN for AD classification in a semi-supervised manner with less annotated T1-MRI images. We remark that the research of GAN adaptation in T1-MRI images is still under development.
 
\section{The Proposed THS-GAN Method}\label{method}

\subsection{Overview}

Fig. \ref{fig_triple_gan} summarizes the architecture of the proposed THS-GAN.
After data preprocessing (see Section \ref{sec-preprocessing}), the normalized T1-MRI images are fed into THS-GAN.
Since the input T1-MRI images are high-order with complicated brain structure, we modify the triple-GAN with the following four significant improvements. (1) Instead of 2D transposed convolution, 3D transposed convolution is utilized in the generator to generate T1-MRI images. (2)  3D-DenseNet \cite{denseNet}\cite{3D-denseNet} is adopted in both the classifier and discriminator to extract subtle features related to AD
within the limited receptive field at a local level. (3) All layers in classifier and discriminator are compressed by Tensor-Train decomposition.
(4) The high-order pooling module GSP block is incorporated into the classifier to make full use of the correlation within feature-maps along the channel axis to capture more discriminative features at the global level to represent the holistic brain. The details of the proposed method will be presented in Section \ref{sec-THS-GAN}.

\subsection{The Architecture}\label{sec-THS-GAN}

The proposed THS-GAN is designed for semi-supervised classification.  Input data $x$ is partially labeled and $y$ represents the corresponding label. $p_{\text{real}}(x)$ denotes the empirical distribution of input data and $p_{\text{real}}(y)$ is assumed as the distribution of labels on partially annotated data. The goal is to predict the label $y$ for both labelled and unlabeled data $x$ as well as to the new generated samples $x$ conditioned on $y$.  As the label $y$ is incomplete, our density model should characterize the uncertainty of both $x$ and $y$, thus the joint distribution $p_{\text{real}}(x,y)$ of image-label pairs can be calculated in two ways: $p_{\text{real}}(x, y)=p_{\text{real}}(y) p_{\text{real}}(x | y)$ and  $p_{\text{real}}(x, y)=p_{\text{real}}(x) p_{\text{real}}(y | x)$. The conditional distributions $p_{\text{real}}(x | y)$ and $p_{\text{real}}(y | x)$ are learnt by the class-conditional generator and auxiliary classifier respectively. Thus the proposed THS-GAN
consists of three networks: (1) a class-conditional generator  that approximately  characterizes the conditional distribution  $p_{g}(x | y) \approx p_{\text{real}}(x | y)$;  (2) a classifier that approximately characterizes the conditional distribution in the opposite  direction $p_{c}(y | x) \approx p_{\text{real}}(y | x)$;  and (3)  a discriminator that distinguishes whether the image-label pair $(x, y)$ comes from the real data distribution $p_{\text{real}}(x, y)$.

More specifically, in the three-player game as illustrated in Fig. \ref{fig_triple_gan}, a sample $x_{\text{unlabel}}$ is drawn from $p_{\text {c}}(x)$, classifier predict label $y_{c}$ given $x_{\text {unlabel}}$ following the conditional distribution $p_{c}(y | x)$. Hence, the pseudo image-label pair $(x_{\text {unlabel}},y_{c})$ is from the joint distribution  $p_{c}(x, y)= p_{c}(x) p_{c}(y | x)$.  Similarly,  a pseudo image-label pair $(x_{g},y_{g})$ is produced by generator given $y_{g}\sim p_{\text {g}}(y)$ by utilizing $x\left|y \sim p_{g}(x | y)\right.$, hence forming the joint distribution $p_{g}(x, y)=p_{g}(y) p_{g}(x | y)$.  With respect to $p_{g}(x | y)$, $x_{g}$ is transformed by generator given label $y_{g}$ and the latent variables $z$.   $x_{g}=G(y_{g}, z)$, $z \sim p_{z}(z)$, where $p_{z}(z)$ is a simple distribution (e.g., uniform or standard normal).   Then the pseudo image-label pairs $(x_{\text{unlabel}}, y_{c})$ and $(x_{g}, y_{g})$  are fed into the discriminator for identification. Discriminator will identify the image-label pairs from real data distribution as positive samples, and discriminator D is trained to maximize the probability of assigning the correct label to both real samples and fake samples from generater G and classifier C. To achieve equilibrium that the joint distributions defined by classifier and   generator both converge to real data distributions, compatible objective of adversarial loss is defined as below:
\begin{equation}
\begin{split}
&    \min _{C, G}    \max _{D} U(C, G, D)   \\   =  &
\mathbb{E}_{(x_{\text {label}}, y) \sim p_{\text {real}}(x, y)}[\log D(x_{\text {label}}, y)]   \\ &
+   \alpha \mathbb{E}_{(x_{\text {unlabel}}, y_{\text {c}}) \sim p_{c}(x, y)}[\log (1-D(x_{\text {unlabel}}, y_{\text {c}}))]    \\
&  +  (1-\alpha) \mathbb{E}_{(x_{g}, y_{g}) \sim p_{g}(x, y)}[\log (1-D(x_{g}, y_{g}))] \\   =  &
\mathbb{E}_{(x_{\text {label}}, y) \sim p_{\text {real}}(x, y)}[\log D(x_{\text {label}}, y)]   \\ &
+  \alpha \mathbb{E}_{x_{\text {unlabel}} \sim p_{\text {c}}(x)}[\log (1-D(x_{\text {unlabel}}, C(x_{\text {unlabel}})))]     \\
&  +   (1-\alpha) \mathbb{E}_{z \sim p_{z}(z),y_{g} \sim p_{\text {g}}(y)}[\log (1-D(G(z,y_{g}), y_{g}))]
\end{split}
\end{equation}
where  $C$, $G$,  and $D$ are individual networks.  $C$ and $D$ are represented by TT-layers (Tensor-Train layers).  $\mathbb{E}_{(x_{\text {label}}, y) \sim p_{\text {real}}(x, y)}$ denotes the expectation over the real labelled data.  $\mathbb{E}_{x_{\text {unlabel}} \sim p_{\text {c}}(x)}$ is the expectation over the real unlabelled data produced by the classifier, and $\mathbb{E}_{\left(x_{g}, y_{g}\right) \sim p_{g}(x, y)}$ is the expectation over the fake data produced by the generator. $D(x_{\text {label}}, y)$ represents the probability that image-label pair came from the real labelled data. Meanwhile,  $D\left(x_{\text {unlabel }}, y_{\mathrm{c}}\right)$ and $D\left(x_{g}, y_{g}\right)$ represent the probability that image-label pair came from fake data produced by classifier and generator respectively. $\alpha \in (0,1)$ is a constant that controls the relative importance of generation and classification, and we use the fixed value of 0.5. The game defined in Equation (1) achieves its equilibrium if and only if $p_{\text {real}}(x, y)=\alpha p_{c}(x, y)  + (1-\alpha) p_{g}(x, y)$. The equilibrium indicates that if one of classifier and generator tends to the real data distribution, the other will also go towards the data distribution, which addresses the competing problem of the conventional semi-supervised GAN. Note that the conventional semi-supervised GAN only contains two players: generator and discriminator. The discriminator shares incompatible roles of identifying fake samples and predicting real labels simultaneously, and the generator estimates the data without considering the labels.  By utilizing the three-player cooperative game,  both the classifier and generator will converge to the real data distribution if the model has been trained to achieve the optimum. In this manner, the class-conditional generator can disentangle different modalities and generate T1-MRI  images to cover all classes (AD, MCI, and  NC).  On the other hand,  the discriminator is trained with dissimilar samples from various classes (AD, MCI, and NC) to provide gradients for the generator. Hence, the mode collapse problem is alleviated.


As aforementioned, layers are tensorized as TT-layer and  we treat the elements of the TT-cores as the parameters of the layer.  TT-layers of classifier and discriminator are represented as various TT-cores $G_{k}$ of elements $\theta_{c}$ and $\theta_{d}$ respectively.
The classifier is updated by descending along its stochastic gradient according to $C_{-} \operatorname{loss}$  with respect to all the elements $\theta_{c}$ of TT-cores. The classifier loss function $C_{-} \operatorname{loss}$ is composed of two parts: the supervised loss and the unsupervised loss:
\begin{equation}
\underbrace{\frac{\partial C_{-} loss}{\partial G_{k}\left[i_{k}, j_{k}\right]}}_{\mathbf{r}_{k-1} \times \mathbf{r}_{k}}    = \nabla_{\theta_{c}}\left[L_{\text {supervised}}+L_{\text {unsupervised}} \right]
\end{equation}
The supervised loss function is defined by the cross-entropy loss of real image-label samples and generated image-label samples in a supervised learning setting:
\begin{equation}
L_{\text {supervised}}  = \mathcal{R}_{\mathcal{L}} + \alpha_{\mathcal{P}} \mathcal{R}_{\mathcal{P}}
\end{equation}
\begin{equation}
\mathcal{R}_{\mathcal{L}}=\mathbb{E}_{(x_{\text {label}}, y) \sim p_{\text {real }}(x, y)}\left[-\log p_{c}(y | x_{\text {label}})\right]
\end{equation}
\begin{equation}
\mathcal{R}_{\mathcal{P}}=\mathbb{E}_{(x_{g}, y_{g}) \sim p_{g}(x, y)}\left[-\log p_{c}(y_{g} | x_{g})\right]
\end{equation}
The cross-entropy loss of real labelled data distribution for classifier is defined as  $\mathcal{R}_{\mathcal{L}}$, which is equivalent to model the KL-divergence between $p_{c}(x, y)$ and $p_{\text {real}}(x, y)$.  As the generated data can also be used for boosting classification performance,  the cross-entropy loss of synthesis data  is defined as $\mathcal{R}_{\mathcal{P}}$, which optimizes classifier on the samples produced by generator in the supervised manner. Minimizing $\mathcal{R}_{\mathcal{P}}$ with respect to classifier is equivalent to minimizing  $D_{K L}\left(p_{g}(x, y) \| p_{c}(x, y)\right)$. Note that directly minimizing   $D_{K L}\left(p_{g}(x, y) \| p_{c}(x, y)\right)$ is infeasible since the unknown likelihood ratio $p_{g}(x, y) / p_{c}(x, y)$  can not be computed directly. $\alpha_{\mathcal{P}}$ is the weight hyperparameter fixed as 0.05.\\
\indent The unsupervised loss is, in fact, the adversarial  loss of standard GAN minimax game: 
\begin{equation}
L_{\text {unsupervised}}   = \mathbb{E}_{x_{\text {unlabel}} \sim p_{\text {c}}(x)}[\log (1-D(x_{\text {unlabel}}, C(x_{\text {unlabel}})))]
\end{equation}
In other words, the unsupervised loss is computed to distinguish real and fake image-label samples. The supervised loss computes the cross-entropy for real classes. In this work, these classes are AD, MCI, and NC.

The generator loss is defined as:
\begin{equation}
	\begin{split}
		G_{-} loss  =    \sum_{\left(x_{g}, y_{g}\right)} \log \left(1-D\left(x_{g}, y_{g}\right)\right)         +    \lambda \left\|x_{\text {label}} - x_{g}\right\|_{L 1}
	\end{split}
\end{equation}
With respect to $G_{-} \operatorname{loss}$ above, a reconstruction loss term is added as the L1 distance between generated images $x_{g}$ and real images $x_{\text {label}}$. $\lambda$ is fixed as 0.01.
The discriminator is updated by descending along its stochastic gradient according to $D_{-} \operatorname{loss}$  with respect to all the elements  $\theta_{d}$  of TT-cores:
\begin{equation}
\begin{split}	
\underbrace{\frac{\partial D_{-} loss}{\partial G_{k}\left[i_{k}, j_{k}\right]}}_{\mathbf{r}_{k-1} \times \mathbf{r}_{k}}     =	& \nabla_{\theta_{d}}  \bigg [ \sum_{\left(x_{\text {label}}, y \right)} \log D\left(x_{\text {label}}, y\right)
\\
&     +    \alpha  \sum_{\left(x_{\text {unlabel}}, y_{c}\right)} \log \left(1-D\left(x_{\text {unlabel}}, y_{c}\right)\right)                \\
&      +   (1-\alpha) \sum_{\left(x_{g}, y_{g}\right)} \log \left(1-D\left(x_{g}, y_{g}\right)\right) \bigg ]
\end{split}
\end{equation}

Intuitively, a sound generator can produce meaningful labeled data beyond training set as auxiliary information for the classifier, which will improve the predictive performance, and vice versa, a sound classifier will boost the performance of the generator. As a result, both the classifier and generator can improve mutually. Moreover, the discriminator can utilize the label information of the unlabeled data through the classifier and then assist the generator to generate correct image-label pairs. Therefore,  THS-GAN is more likely to reach Nash equilibrium.

Two components of triple-GAN (classifier and discriminator) are converted to the Tensor-Train format (TT-format) \cite{novikov15tensornet,garipov16ttconvtensornet2,28tensor}. We refer to 1-D data as a vector, denoted as $v$. 2-D array is matrix, denoted as $\boldsymbol{V}$, and higher dimensional array is tensor, denoted as $\mathcal{V}$. To refer one specific element from a tensor, we use $\mathcal{V}(i)=\mathcal{V}\left(i_{1}, i_{2}, \ldots i_{d}\right)$, where d is the dimensionality of the tensor $\mathcal{V}$ and $i$ is the index vector.  Our proposed THS-GAN ingests T1-MRI image as 3D tensor, where each dimension corresponds to height, width, and slice respectively.  A $d$-dimensional $n_{1} \times n_{2} \times \ldots \times n_{d}$ tensor $\mathcal{V}$ can be represented in the TT-format\cite{layerTensor} \cite{28tensor} as:
\begin{equation}\label{equ_tensor}
\mathcal{V}\left(i_{1}, i_{2}, \ldots i_{d}\right)=G_{1}\left[i_{1}\right] G_{2}\left[i_{2}\right] \ldots G_{d}\left[i_{d}\right]
\end{equation}
where $G_{k}\left[i_{k}\right]$ is an $r_{k-1} \times r_{k}$ matrix, which is one slice from the 3-dimensional array $G_{k}$. The elements of the collection $\left\{r_{k}\right\}_{k=0}^{d}$ are called TT-ranks.  $r_{0}=r_{d}=1$ is the boundary condition to keep the matrix product (\ref{equ_tensor}) of size $1 \times 1$.

The collections of matrices $\left\{\left\{G_{k}\left[j_{k}\right]\right\}_{j_{k}=1}^{n_{k}}\right\}_{k=1}^{d}$ are called TT-cores\cite{garipov16ttconvtensornet2}.  The TT-format requires $\sum_{k=1}^{d} n_{k} r_{k-1} r_{k}$ parameters to represent a tensor $\mathcal{V} \in \mathbb{R}^{n_{1} \times \ldots \times n_{d}}$ which has $\prod_{k=1}^{d} n_{k}$ elements. The TT-ranks $r_{k}$ control the trade-off between the number of parameters and the accuracy of the representation. The smaller the TT-ranks, the more memory efficient the TT-format is. But if the TT-ranks are set too small, the accuracy might deteriorate due to information loss caused by over-compressing.        Such a representation is memory-efficient to store high-order data. Meanwhile,  the significant structural information of data can be preserved.  These properties are suitable for representing T1-MRI images.
In the following, we introduce tensor-train decomposition for fully-connected layers and convolutional layers respectively.
\begin{figure*}[ht] 
	\includegraphics[width=\linewidth]{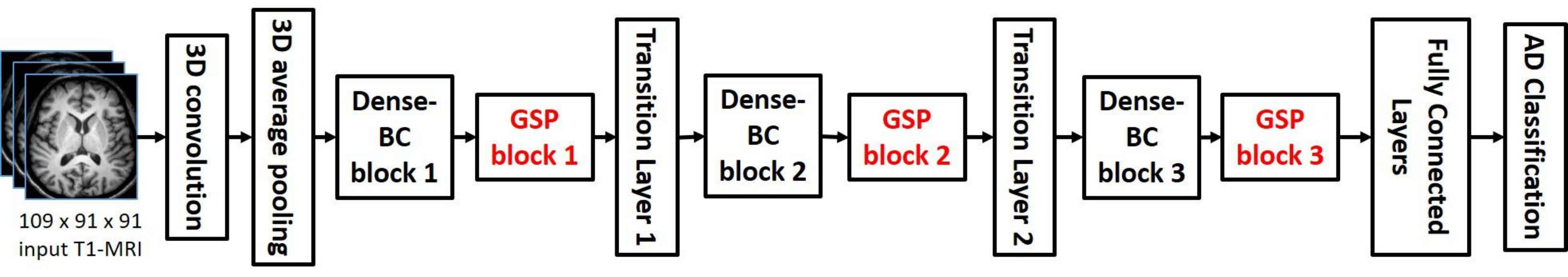}
	\caption{The classifier framework is composed of 3D-DenseNet and GSP block.  Note that only one GSP block is inserted at one of three optional positions in red. There is no GSP block in discriminator.}
	\label{fig_classifierFramework}
\end{figure*}
\subsubsection{Fully-Connected Layers Tensor-train Decomposition }

The fully-connected layer is applied to an input N-dimensional vector $X$:
\begin{equation}\label{equ_fc}
Y=\boldsymbol{W}X+B
\end{equation}
where the weight matrix $\boldsymbol{W} \in \mathbb{R}^{M \times N}$ and the bias vector $B \in \mathbb{R}^{M}$ define the linear transformation.
A TT-fully-connected-layer transforms a d-dimensional tensor $\mathcal{X}$ (which is constructed from the corresponding vector $X$) to the d-dimensional tensor $\mathcal{Y}$ (which corresponds to the output vector $Y$) by factorizing the weight matrix $\boldsymbol{W}$ into the TT-format with the TT-cores $G_{k}\left[i_{k}, j_{k}\right]$. Thus the linear transformation (Equation (\ref{equ_fc})) of a fully-connected layer can be represented in the TT-layer:
\begin{equation}\label{equ_fc_tensor}
\begin{split}
\mathcal{Y}\left(i_{1}, \ldots, i_{d}\right)= 	& \sum_{j_{1}, \ldots, j_{d}} G_{1}\left[i_{1}, j_{1}\right] \ldots G_{d}\left[i_{d}, j_{d}\right] \mathcal{X}\left(j_{1}, \ldots, j_{d}\right) \\
&+  \mathcal{B}\left(i_{1}, \ldots, i_{d}\right),
\end{split}
\end{equation}
where   $G\left[i_{d}, j_{d}\right] \in \mathbb{R}^{r_{k-1} \times r_{k}}$ is a slice of cores as illustrated in the red part of
Fig. \ref{fig_triple_gan}.  Since the fully-connected layer is a special case of convolutional layer with kernel size $1 \times 1\times 1$, such TT-format can also be applied to convolutional layers in a similar manner.

\subsubsection{Convolutional layers Tensor-train Decomposition}

3D convolution is an extension of 2D convolution with one more spatial dimension in terms of slice with respect to T1-MRI volume. The traditional 3D convolutional layer transforms the 4-dimensional input tensor $\mathcal{X} \in \mathbb{R}^{W \times H \times L \times C}$ into the output $\mathcal{Y} \in \mathbb{R}^{W^{\prime} \times H^{\prime} \times L^{\prime} \times S}$ by convolving $\mathcal{X}$ with the kernel $\mathcal{K} \in \mathbb{R}^{\ell \times \ell \times \ell \times C \times S}$:
\begin{equation}\label{equ_4Dconv}
\begin{split}
\mathcal{Y}(x, y,z, s)=& \sum_{i=1}^{\ell} \sum_{j=1}^{\ell} \sum_{k=1}^{\ell}  \sum_{c=1}^{C} \mathcal{K}(i, j,k,c,s)  \\
&\mathcal{X}(x+i-1, y+j-1, z+k-1, c)
\end{split}
\end{equation}
When stride is set as 1 and there is no zero padding, $W^{\prime}= W-l+1$, $H^{\prime}= H-l+1$ and $L^{\prime}= L-l+1$. The Tensor-Train decomposition is applied to the convolutional kernel $\mathcal{K}$ as follows:
\begin{equation}\label{equ_kernel}
\mathcal{K}\left(x, y,z,c,s\right)=G_{0}[i, j,k] G_{1}\left[c_{1}, s_{1}\right] \ldots G_{d}\left[c_{d}, s_{d}\right]
\end{equation}
Red part of Fig. \ref{fig_triple_gan} also presents an illustration for Equation (\ref{equ_kernel}), and the 3D convolutional layer is converted to TT-layer as follows:
\begin{equation}
\mathcal{X}(x, y,z, c) \stackrel{\text { reshape }}{\longrightarrow} \tilde{\mathcal{X}}\left(x, y,z, c_{1}, c_{2}, \ldots, c_{d}\right),
\end{equation}
\begin{equation}
\mathcal{Y}(x, y,z, s) \stackrel{\text { reshape }}{\longrightarrow} \widetilde{\mathcal{Y}}\left(x, y, z,s_{1}, s_{2}, \ldots, s_{d}\right),
\end{equation}
and
\begin{equation}\label{equ_tensor_4Dconv}
\begin{split}
&\tilde{\mathcal{Y}}\left(x, y,z, s_{1}, \ldots, s_{d}\right) \\    = &\sum_{i=1}^{\ell} \sum_{j=1}^{\ell} \sum_{k=1}^{\ell} \sum_{c_{1}, \ldots, c_{d}}  G_{0}[i, j,k] G_{1}\left[c_{1}, s_{1}\right] \ldots G_{d}\left[c_{d}, s_{d}\right]     \\ &   \widetilde{\mathcal{X}}\left(i+x-1, j+y-1,k+z-1, c_{1}, \ldots, c_{d}\right), \end{split}
\end{equation}
where $c=\prod_{i=1}^{d} c_{i}$, $s=\prod_{i=1}^{d} s_{i}$ and $d$ is the number of TT-cores.
By replacing the 4D convolutional kernel with approximations using lower rank matrices, redundancy in convolutional layers can be removed implicitly.  It is worth noting that although applying tensor-train decomposition to neural networks can achieve a large factor of compression, finding optimal TT-ranks remains difficult \cite{novikov15tensornet} \cite{TGAN}. The TT-layer is compatible with the existing training algorithms for neural networks because all the derivatives required by the back-propagation algorithm can be computed using the properties of the TT-format.

The network utilized in both the classifier and discriminator is DenseNet \cite{denseNet}. We expand it to 3D-DenseNet by adding a spatial dimension to all convolutional and pooling layers in DenseNet for 3D T1-MRI volume. Feature-maps learned by all preceding layers are concatenating along the last dimension for the subsequent layers. Through such dense connectivity, feature-maps are reused and the vanishing-gradient problem is alleviated. Meanwhile, 3D-DenseNet can extract the local morphological features related to AD lesions from the whole volumes efficiently. The details of 3D-denseNet are found in \cite{denseNet}\cite{3D-denseNet}. In this paper, the depth is set as 30, the growth rate is set as 12, the number of the Dense-BC block is set as 3, and reduction is set as 0.5.

Furthermore, the high-order pooling module GSP block can make full use of the second-order statistics of the holistic MRI images.  The long-range dependencies between slices of different directions can be effectively captured  for extracting more significant features in a self-attention manner. Thus the GSP block is added after the Dense-BC block in the classifier, as illustrated in Fig. \ref{fig_classifierFramework}, aiming to learn more discriminative representations by re-calibrating the 4D channel-wise feature-maps. There is one more GSP block (in red), which can be positioned at: (1) GSP block 1, (2) GSP block 2, or (3) GSP block 3.

\begin{figure*} 
	\includegraphics[width=\linewidth]{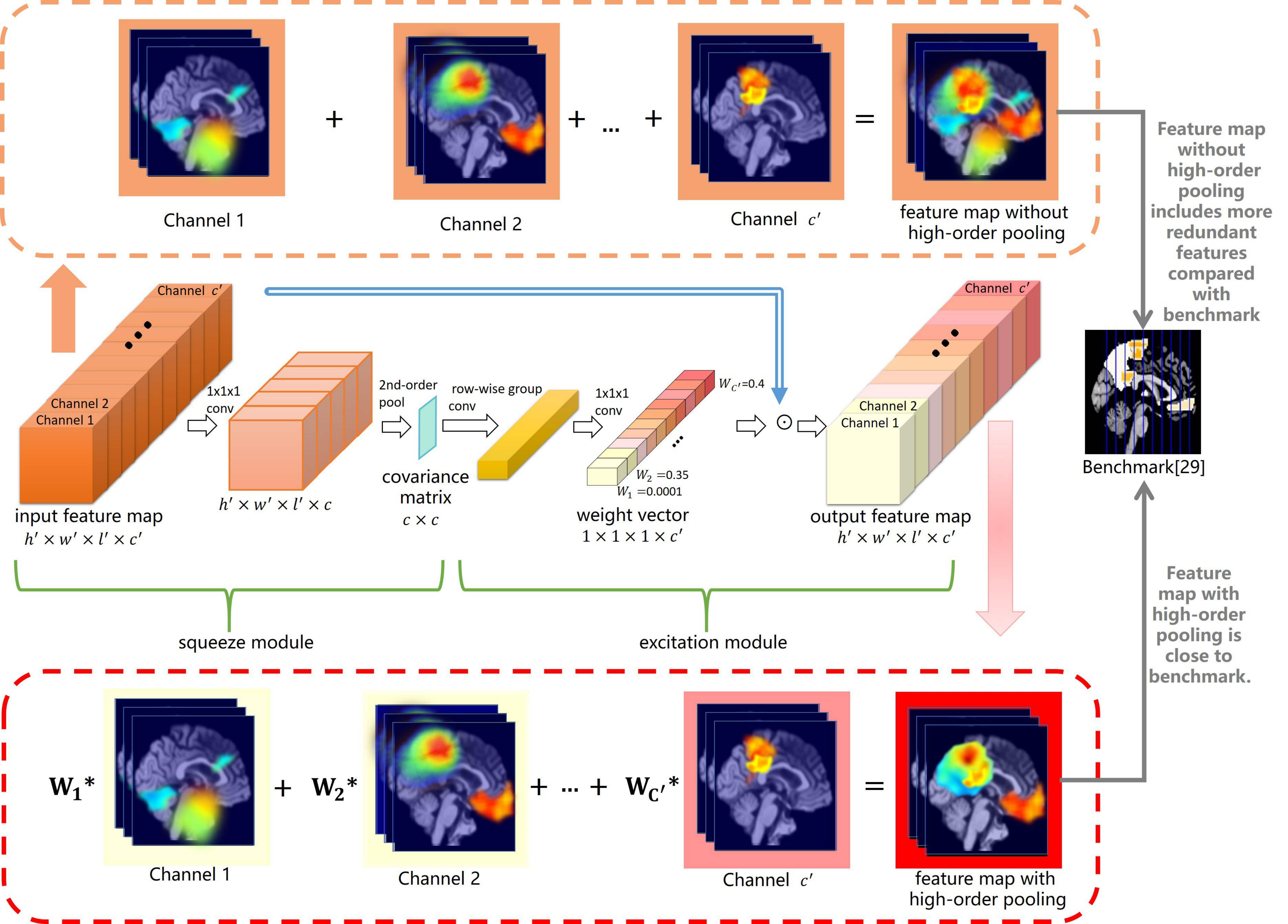}
	\caption{ High-order pooling module GSP block. Given an input 4D feature map, $1 \times 1 \times 1$ convolution is performed to reduce dimension. Then the covariance matrix is computed followed by convolution and non-linear activation, finally weight vector is produced to recalibrate the feature map along the channel dimension. The high-order pooling can capture the dependency of features at distant positions by computing all pairwise channel correlations. As a result, significant features will be enhanced.  As each channel corresponds to a particular feature, each feature map of all channels is considered as a feature set that can map back to individual voxels of input MRI image. The discriminative features related to AD are shown in the benchmark\cite{ADroi}. }
	\label{fig_GSP_block}
\end{figure*}

Inspired by \cite{GSP}, the GSP block is extended to a 4D tensor, as illustrated in Fig. \ref{fig_GSP_block}. Given a 4D feature map outputted by a previous Dense-BC block, we first perform GSP to model pairwise channel correlations of the holistic feature map. Then the resulting covariance matrix is processed by convolutions and non-linear activations, which is finally used for scaling the 4D feature map  along the channel dimension.

More specifically, the GSP block consists of two modules: a squeeze module and an excitation module.   The squeeze module aims to model the second-order statistics along the channel dimension of the input  feature map  for capturing channel dependency.  Consider a 4D feature map of $h^{\prime} \times w^{\prime} \times l^{\prime} \times c^{\prime}$  as an input, where $h^{\prime}$ is the spatial height of the feature-map,  $w^{\prime}$ is the width, $l^{\prime}$ is depth, and $c^{\prime}$ is the number of channels. It  can be seen as  $c^{\prime}$ cubes where each cube is of size $h^{\prime} \times w^{\prime} \times l^{\prime}$.   First,  $1 \times 1 \times 1$ convolution is utilized to reduce the number of channels from $c^{\prime}$ to $c$ ($c<c^{\prime}$) to decrease the computational cost of the following operations. For the $h^{\prime} \times w^{\prime} \times l^{\prime} \times c$ tensor of reduced dimensionality, the pairwise channel correlations are computed to one $c \times c$ covariance matrix.   The resulting covariance matrix has clear physical meaning, its $i^{th}$  row indicates the statistical dependency of channel $i$  with all channels. As the quadratic operations  involved change the order of data, row-wise normalization is performed for the covariance matrix with respect to the  structural information of brain.  To simplify the block design and to find the appropriate trade-off between computational complexity and classification accuracy, we calculate the size of the covariance matrix as $c=c^{\prime}/6$ in a self-adaptive manner.

The excitation module aims to scale the channel for feature re-calibration.  In the excitation module, before channel scaling, we perform two consecutive operations of convolution and non-linear activation for the covariance matrix. To maintain the structural information, the covariance matrix is processed with row-wise convolution, which is followed by a Leaky Rectified Linear Unit (LReLU). Then we perform the second convolution and the sigmoid function as a non-linear activation to compute the weight vector of [$\mathbf{W}_{1}$,$\mathbf{W}_{2}$,...,$\mathbf{W}_{\mathrm{c}^{\prime}}$]. The final output of the GSP block
is obtained by operating the dot product between the weight vector [$\mathbf{W}_{1}$,$\mathbf{W}_{2}$,...,$\mathbf{W}_{\mathrm{c}^{\prime}}$] and the respective channels [Channel 1,Channel 2,...,Channel $\mathrm{c}^{\prime}$].  Individual channels are thus emphasized or suppressed in this soft manner in terms of the weights. Thus the discriminative features related to AD lesions are enhanced, and redundant features are suppressed. As shown in  Fig. \ref{fig_GSP_block}, the feature map output by GSP block is close to the benchmark with less redundant features, and all significant features are discovered. On the other hand, the feature map without high-order pooling includes more redundant features compared with the benchmark.

Furthermore, the network structure of each component in THS-GAN is further optimized from the following perspectives.
(1) For generator, the condition variable $y$ can either be concatenated with the random noise $z$ in the first layer or be added in the subsequent layers as additional channels. In our study, we adopt the latter one.  (2) As suggested by Radford et al. \cite{DCGAN}, we also add Batch Normalization (BN) to both the discriminator and the generator in the THS-GAN model to prevent the generator from collapsing all the samples to a single point. However, adding BN to all layers causes model instabilities. Hence we also avoid using BN in the generator output layer and the discriminator input layer as they suggest. The tanh function is used in the generator output layer. (3) The first order pooling (average pooling) is still utilized since the GSP block can not reduce dimensions of the feature-map resulting in a large number of parameters. Thus the first order pooling is combined with GSP block to abstract the discriminative representations, so that the proposed THS-GAN model can take advantage of both first-order and second-order statistics for AD diagnosis.

\section{Experiments and Results}\label{experimental_results}

\subsection{Dataset and Preprocessing}\label{sec-preprocessing}

A total of 833 T1-weighted MRI images are downloaded from  ADNI\footnote[1]{ http://adni.loni.usc.edu/}  database in the neuroimaging informatics technology initiative (NIfTI) format, which have already been processed for spatial distortion correction caused by gradient nonlinearity and B1 field inhomogeneity.   The standard image pre-processing procedure is performed on the selected T1-MRI images for each subject, including grad-warping, skull-stripping, cerebellum removal, and intensity correction. We perform skull stripping using Brain Extraction Tool (FSL-BET), followed by a manual correction to ensure that both skull and dura have been removed completely. Then, we remove the cerebellum by warping a labeled template to each skull-stripped image.   Finally, all brain images were aligned to the standardized MIN152  template using FSL FLIRT \cite{flirt}.     The dimension of each image is $109 \times  91 \times 91$. Each image comprises 109 2D slices of $91 \times  91$.

Among 833 T1-MRI images, there are 221 AD subjects, 297 MCI subjects, and 315 Normal Controls (NC) subjects respectively. To evaluate the
effectiveness of our model, we set up three groups of experiments: (1) AD vs. NC,  (2) MCI vs. NC, and (3) AD vs. MCI classification.  It is worth noting that the second classification is significant to distinguish MCI from NC for early diagnosis so that timely therapeutic interventions can be carried out to slow down the progression of MCI to AD.

The T1-MRI image is normalized into the range [-1,1], and the whole volume of $109 \times 91 \times 91$ voxels is fed into the proposed THS-GAN model as a tensor directly without compressing or downsizing to ensure no information loss. No data augmentation was used. For evaluation, 10\% of the total data is selected as a validation dataset and another 10\%  as a test dataset. The remainding (80\%) was used as a training dataset for our
THS-GAN model. The validation dataset was utilized to tune hyperparameters to find the best model out of several trained models.

\subsection{Experimental Setup}

The proposed THS-GAN model is trained on the ADNI dataset from scratch in an end-to-end manner. We implement our method based on  TensorFlow\footnote[1]{http://www.tensorflow.org/}. It takes around 10 hours for training our model on the training data set along with validation on the validation set at each epoch on NVIDIA GeForce GTX 1080 GPU. The initial learning rate is 0.01 and will decrease to $10^{-3}$ at 75 epochs and $10^{-4}$ at 110 epochs.  A weight decay of $10^{-4}$ is applied for all the weights, and we use stochastic gradient descent with  Nesterov momentum \cite{Momentum} of coefficient 0.9. The validation accuracy will be evaluated once for each training epoch. Besides, we set the batch size of both labeled data and unlabelled data as 7, and the number of epochs as 150.  The loss $\mathcal{R}_{\mathcal{P}}$  is not applied until the number of epochs reaches a threshold that the generator can generate meaningful data. We search the threshold in \{60,120\} based on the validation performance, and $\alpha_{\mathcal{P}}$ is fixed as 0.05.

\subsection{Evaluation Metrics}

Five metrics are used for quantitative evaluation and comparison, including accuracy, precision, recall, f1-score, and AUC.  The Area Under a ROC curve (AUC) is a single value frequently used to measure classifier performance ($0 \leq  AUC  \leq 1$). In other words, AUC is an indicator of the probability that a classifier will correctly classify instances. Note that an AUC value of 0.5 indicates a random classifier (guessing). We denote TP, TN, FP, and FN as true positive, true negative, false positive, and false negative respectively. The evaluation metrics are defined as follows:
\begin{equation}\label{eq:precision}
\text{Precision}=\frac{TP}{TP+FP},
\end{equation}
\begin{equation}\label{eq:recall}
\text{Recall}=\frac{TP}{TP+FN},
\end{equation}
\begin{equation}\label{eq:f1Score}
\text{F1-score}=2 \times \frac{\text {Precision} \times \text {Recall}}{\text {Precision}+\text {Recall}},
\end{equation}
and
\begin{equation}\label{eq:accuracy}
\text{Accuracy}=\frac{TP+TN}{TP+TN+FP+FN}.
\end{equation}

\subsection{The effect of TT-core number} \label{sec_coreNumber}

\begin{figure}
	\includegraphics[width=\linewidth]{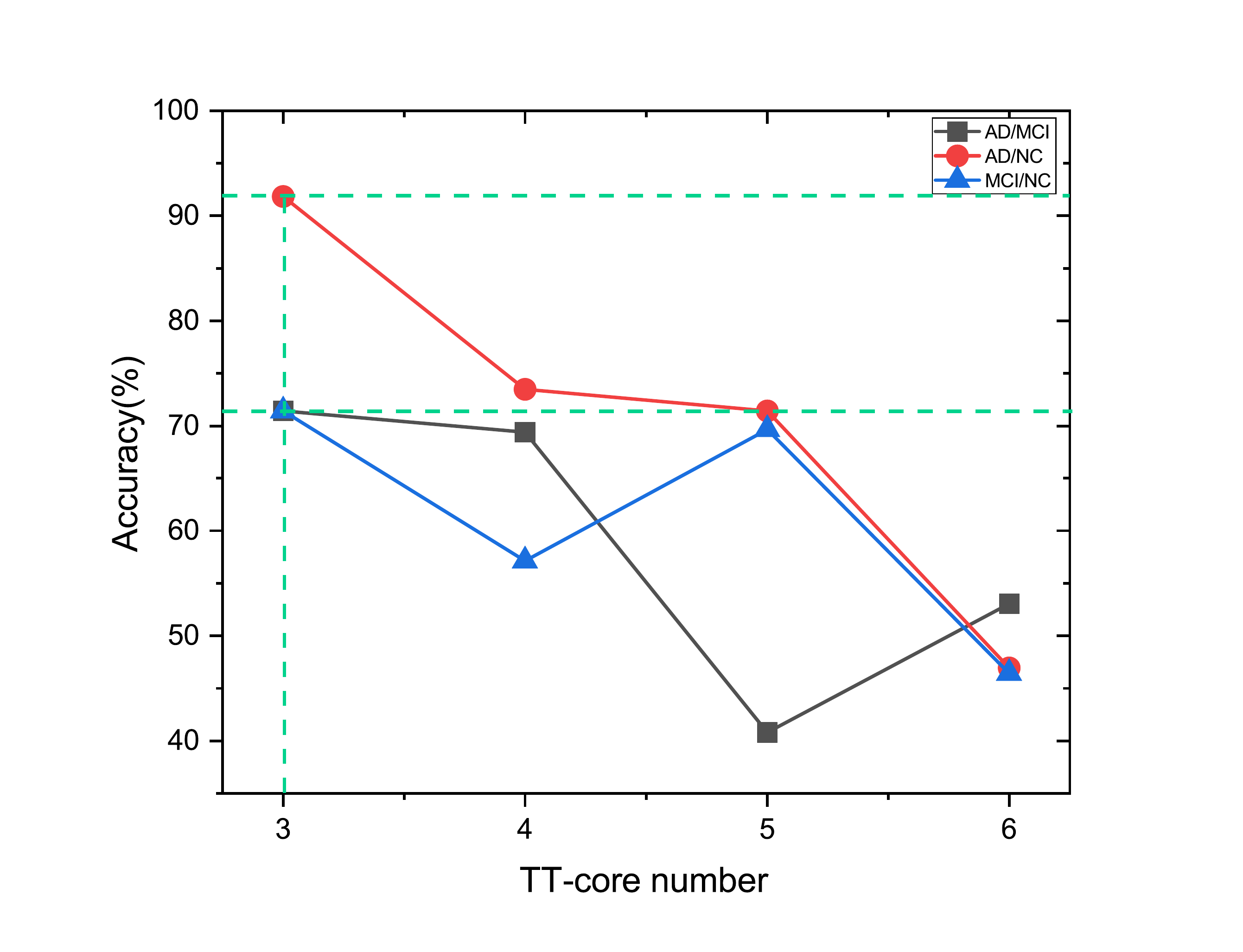}
	\caption{Comparison of different TT-core numbers.}
	\label{fig_core_Number}
\end{figure}

As mentioned in Section \ref{sec-THS-GAN}, the TT-core number and the TT-rank are two parameters that have a great impact on classification results. This section provides a comparative evaluation of the proposed THS-GAN with respect to a range of TT-core numbers. The GSP block is fixed at the position of  ``GSP block 3''. TT-rank of the classifier and discriminator was fixed at  14 and 6 respectively.  Fig. \ref{fig_core_Number} shows that as the TT-core number increased from 3 to 6, the classification accuracy decreased for AD/NC classification. Meanwhile, for AD/MCI and MCI/NC classification, there are no specific trends of accuracy as core number increased from 3 to 6.
But similar to AD/NC classification, the best accuracy is achieved at the minimal core number. This observation is consistent with \cite{novikov15tensornet}. Thus we set the TT-core number as 3 in the rest of the experiments.

\subsection{The effect of TT-rank and GSP block position}\label{experimentTTrank}

\begin{table*}[htbp]
	\centering
	\caption{Comparison of THS-GAN using different GSP block positions and TT-ranks for AD/NC classification.}
	\begin{tabular}{|c|c|c|c|c|c|c|c|c|c|}
		\hline
		GSP block Position &   C\_rank & D\_rank & \#parameters & AUC(\%) & Accuracy(\%) & Class & precision(\%) & recall(\%) & f1-score(\%) \\ \hline
		\multirow{14}{*}{GSP block 1} &   \multirow{2}{*}{14} & \multirow{2}{*}{6} & \multirow{2}{*}{118,210} & \multirow{2}{*}{50.00} & \multirow{2}{*}{56.00} & AD & 56.00 & 100 & 71.79 \\ \cline{7-10}
		&  &  &  &  &     & NC & 0 & 0 & 0 \\ \cline{2-10}
		& \multirow{2}{*}{15} & \multirow{2}{*}{7} & \multirow{2}{*}{139,611} & \multirow{2}{*}{50.00} & \multirow{2}{*}{44.00} & AD & 44.00 & 100 & 61.11 \\ \cline{7-10}
		&  &  &    &  &  & NC & 0 & 0 & 0 \\ \cline{2-10}
		& \multirow{2}{*}{16} & \multirow{2}{*}{8} & \multirow{2}{*}{163,516} & \multirow{2}{*}{50.00} & \multirow{2}{*}{60.00} & AD & 60.00 & 100 & 75.00 \\ \cline{7-10}
		&  &  &   &  &  & NC & 0 & 0 & 0 \\ \cline{2-10}
		& \multirow{2}{*}{17} & \multirow{2}{*}{9} & \multirow{2}{*}{189,925} & \multirow{2}{*}{84.00} & \multirow{2}{*}{84.00} & AD & 84.00 & 84.00 & 84.00 \\ \cline{7-10}
		&  &   &  &  &  & NC & 84.00 & 84.00 & 84.00 \\ \cline{2-10}
		& \multirow{2}{*}{18} & \multirow{2}{*}{10} & \multirow{2}{*}{218,838} & \multirow{2}{*}{63.33} & \multirow{2}{*}{77.55} & AD & 100 & 26.67 & 42.11 \\ \cline{7-10}
		&  &  &  &   &  & NC & 75.56 & 100 & 86.08 \\ \cline{2-10}
		& \multirow{2}{*}{19} & \multirow{2}{*}{11} & \multirow{2}{*}{250,255} & \multirow{2}{*}{69.64} & \multirow{2}{*}{62.22} & AD & 100 & 39.29 & 56.41 \\ \cline{7-10}
		&  &   &  &  &  & NC & 50.00 & 100 & 66.67 \\ \cline{2-10}
		& \multirow{2}{*}{\textit{20}} & \multirow{2}{*}{\textit{12}} & \multirow{2}{*}{284,176} & \multirow{2}{*}{\textit{91.99}} & \multirow{2}{*}{\textit{92.00}} & AD & 92.31 & 92.31 & 92.31 \\ \cline{7-10}
		&  &  &  &  &    & NC & 91.67 & 91.67 & 91.67 \\ \hline
		\multirow{14}{*}{\textbf{GSP block 2}} &  \multirow{2}{*}{14} & \multirow{2}{*}{6} & \multirow{2}{*}{120,034} & \multirow{2}{*}{83.33} & \multirow{2}{*}{79.59} & AD & 100 & 66.67 & 80.00 \\ \cline{7-10}
		&  &  &  &    &  & NC & 65.52 & 100 & 79.17 \\ \cline{2-10}
		& \multirow{2}{*}{15} & \multirow{2}{*}{7} & \multirow{2}{*}{141,435} & \multirow{2}{*}{50.00} & \multirow{2}{*}{48.98} & AD & 0 & 0 & 0 \\ \cline{7-10}
		&  &  &    &  &  & NC & 48.98 & 100 & 65.75 \\ \cline{2-10}
		& \multirow{2}{*}{16} & \multirow{2}{*}{8} & \multirow{2}{*}{165,340} & \multirow{2}{*}{93.18} & \multirow{2}{*}{93.88} & AD & 100 & 86.36 & 92.68 \\ \cline{7-10}
		&  &  &    &  &  & NC & 90.00 & 100 & 94.74 \\ \cline{2-10}
		& \multirow{2}{*}{17} & \multirow{2}{*}{9} & \multirow{2}{*}{191,749} & \multirow{2}{*}{44.71} & \multirow{2}{*}{55.10} & AD & 60.47 & 83.87 & 70.27 \\ \cline{7-10}
		&  &    &  &  &  & NC & 16.67 & 5.56 & 8.34 \\ \cline{2-10}
		& \multirow{2}{*}{18} & \multirow{2}{*}{10} & \multirow{2}{*}{220,662} & \multirow{2}{*}{83.36} & \multirow{2}{*}{83.67} & AD & 85.71 & 78.26 & 81.82 \\ \cline{7-10}
		&  &  &    &  &  & NC & 82.14 & 88.46 & 85.18 \\ \cline{2-10}
		& \multirow{2}{*}{19} & \multirow{2}{*}{11} & \multirow{2}{*}{252,079} & \multirow{2}{*}{83.88} & \multirow{2}{*}{83.67} & AD & 88.89 & 82.76 & 85.72 \\ \cline{7-10}
		&  &     &  &  &  & NC & 77.27 & 85.00 & 80.95 \\ \cline{2-10}
		& \multirow{2}{*}{\textbf{20}} & \multirow{2}{*}{\textbf{12}} & \multirow{2}{*}{\textbf{286,000}} & \multirow{2}{*}{\textbf{95.92}} & \multirow{2}{*}{\textbf{95.92}} & AD & \textbf{95.83} & \textbf{95.83} & \textbf{95.83} \\ \cline{7-10}
		&  &  &  &    &  & NC & \textbf{96.00} & \textbf{96.00} & \textbf{96.00} \\ \hline
		\multirow{14}{*}{GSP block 3} &  \multirow{2}{*}{14} & \multirow{2}{*}{6} & \multirow{2}{*}{121,048} & \multirow{2}{*}{91.81} & \multirow{2}{*}{91.84} & AD & 91.30 & 91.30 & 91.30 \\ \cline{7-10}
		&  &  &  &    &  & NC & 92.31 & 92.31 & 92.31 \\ \cline{2-10}
		& \multirow{2}{*}{15} & \multirow{2}{*}{7} & \multirow{2}{*}{142,449} & \multirow{2}{*}{74.48} & \multirow{2}{*}{73.47} & AD & 83.33 & 68.97 & 75.47 \\ \cline{7-10}
		&  &  &   &  &  & NC & 64.00 & 80.00 & 71.11 \\ \cline{2-10}
		& \multirow{2}{*}{16} & \multirow{2}{*}{8} & \multirow{2}{*}{166,354} & \multirow{2}{*}{84.32} & \multirow{2}{*}{81.63} & AD & 95.83 & 74.19 & 83.63 \\ \cline{7-10}
		&  &  &  &   &  & NC & 68.00 & 94.44 & 79.07 \\ \cline{2-10}
		& \multirow{2}{*}{17} & \multirow{2}{*}{9} & \multirow{2}{*}{192,763} & \multirow{2}{*}{80.77} & \multirow{2}{*}{79.59} & AD & 69.70 & 100 & 82.14 \\ \cline{7-10}
		&  &  &   &  &  & NC & 100 & 61.54 & 76.19 \\ \cline{2-10}
		& \multirow{2}{*}{18} & \multirow{2}{*}{10} & \multirow{2}{*}{221,676} & \multirow{2}{*}{73.82} & \multirow{2}{*}{73.47} & AD & 79.17 & 70.37 & 74.51 \\ \cline{7-10}
		&  &  &  &  &  & NC & 68.00 & 77.27 & 72.34 \\ \cline{2-10}
		& \multirow{2}{*}{19} & \multirow{2}{*}{11} & \multirow{2}{*}{253,093} & \multirow{2}{*}{69.40} & \multirow{2}{*}{69.39} & AD & 72.00 & 69.23 & 70.59 \\ \cline{7-10}
		&  &  &  &  &  & NC & 66.67 & 69.57 & 68.09 \\ \cline{2-10}
		& \multirow{2}{*}{\textit{20}} & \multirow{2}{*}{\textit{12}} & \multirow{2}{*}{287,014} & \multirow{2}{*}{\textit{92.00}} & \multirow{2}{*}{\textit{91.84}} & AD & 85.71 & 100 & 92.31 \\ \cline{7-10}
		&  &  &  &    &  & NC & 100 & 84.00 & 91.30 \\ \hline
		\multicolumn{3}{|l|}{\multirow{2}{*}{SS-GAN \cite{SSgan} }} & \multirow{2}{*}{251,637} & \multirow{2}{*}{80.02} & \multirow{2}{*}{80.39} & AD & 82.76 & 82.76 & 82.76 \\ \cline{7-10}
		\multicolumn{3}{|l|}{} &  &  &  & NC & 77.27 & 77.27 & 77.27 \\ \hline
		\multicolumn{3}{|l|}{\multirow{2}{*}{triple-GAN \cite{tripleGAN} }} & \multirow{2}{*}{506,386} & \multirow{2}{*}{86.83} & \multirow{2}{*}{87.76} & AD & 90.32 & 90.32 & 90.32 \\ \cline{7-10}
		\multicolumn{3}{|l|}{} &  &  &  & NC & 83.33 & 83.33 & 83.33 \\ \hline
	\end{tabular}
	\label{table_TTrank_accuracy_AD_NC}
\end{table*}

\begin{table*}[htbp]
	\centering
	\caption{Comparison of THS-GAN using different GSP block positions and TT-ranks for MCI/NC classification.}
	\begin{tabular}{|c|c|c|c|c|c|c|c|c|c|}
		\hline
		GSP block Position &  C\_rank & D\_rank & \#parameters & AUC(\%) & Accuracy(\%) & Class & precision(\%) & recall(\%) & f1-score(\%) \\ \hline
		\multirow{14}{*}{GSP block 1} &  \multirow{2}{*}{14} & \multirow{2}{*}{6} & \multirow{2}{*}{118,210} & \multirow{2}{*}{70.13} & \multirow{2}{*}{74.07} & MCI & 69.77 & 96.77 & 81.08 \\ \cline{7-10}
		&  &  &  &  &     & NC & 90.91 & 43.48 & 58.83 \\ \cline{2-10}
		& \multirow{2}{*}{15} & \multirow{2}{*}{7} & \multirow{2}{*}{139,611} & \multirow{2}{*}{84.89} & \multirow{2}{*}{85.19} & MCI & 81.25 & 92.86 & 86.67 \\ \cline{7-10}
		&  &  &  &    &  & NC & 90.91 & 76.92 & 83.33 \\ \cline{2-10}
		& \multirow{2}{*}{16} & \multirow{2}{*}{8} & \multirow{2}{*}{163,516} & \multirow{2}{*}{64.94} & \multirow{2}{*}{59.26} & MCI & 48.72 & 90.48 & 63.34 \\ \cline{7-10}
		&  &  &  &    &  & NC & 86.67 & 39.39 & 54.16 \\ \cline{2-10}
		& \multirow{2}{*}{17} & \multirow{2}{*}{9} & \multirow{2}{*}{189,925} & \multirow{2}{*}{79.94} & \multirow{2}{*}{81.48} & MCI & 84.21 & 69.57 & 76.19 \\ \cline{7-10}
		&  &  &  &    &  & NC & 80.00 & 90.32 & 84.85 \\ \cline{2-10}
		& \multirow{2}{*}{\textit{18}} & \multirow{2}{*}{\textit{10}} & \multirow{2}{*}{218,838} & \multirow{2}{*}{\textit{85.71}} & \multirow{2}{*}{\textit{85.19}} & MCI & 100 & 71.43 & 83.33 \\ \cline{7-10}
		&  &  &    &  &  & NC & 76.47 & 100 & 86.67 \\ \cline{2-10}
		& \multirow{2}{*}{19} & \multirow{2}{*}{11} & \multirow{2}{*}{250,255} & \multirow{2}{*}{71.56} & \multirow{2}{*}{70.37} & MCI & 62.16 & 92.00 & 74.19 \\ \cline{7-10}
		&  &  &    &  &  & NC & 88.24 & 51.72 & 65.22 \\ \cline{2-10}
		& \multirow{2}{*}{20} & \multirow{2}{*}{12} & \multirow{2}{*}{284,176} & \multirow{2}{*}{66.48} & \multirow{2}{*}{61.11} & MCI & 51.22 & 95.45 & 66.67 \\ \cline{7-10}
		&   &  &  &  &  & NC & 92.31 & 37.50 & 53.33 \\ \hline
		\multirow{14}{*}{GSP block 2} &   \multirow{2}{*}{14} & \multirow{2}{*}{6} & \multirow{2}{*}{120,034} & \multirow{2}{*}{65.74} & \multirow{2}{*}{62.50} & MCI & 54.55 & 96.00 & 69.57 \\ \cline{7-10}
		&  &  &  &  &     & NC & 91.67 & 35.48 & 51.16 \\ \cline{2-10}
		& \multirow{2}{*}{\textit{15}} & \multirow{2}{*}{\textit{7}} & \multirow{2}{*}{141,435} & \multirow{2}{*}{\textit{88.32}} & \multirow{2}{*}{\textit{87.50}} & MCI & 96.15 & 80.65 & 87.72 \\ \cline{7-10}
		&  &  &  &  &     & NC & 80.00 & 96.00 & 87.27 \\ \cline{2-10}
		& \multirow{2}{*}{16} & \multirow{2}{*}{8} & \multirow{2}{*}{165,340} & \multirow{2}{*}{52.23} & \multirow{2}{*}{53.57} & MCI & 57.14 & 14.81 & 23.52 \\ \cline{7-10}
		&  &  &  &    &  & NC & 53.06 & 89.66 & 66.67 \\ \cline{2-10}
		& \multirow{2}{*}{17} & \multirow{2}{*}{9} & \multirow{2}{*}{191,749} & \multirow{2}{*}{70.18} & \multirow{2}{*}{69.64} & MCI & 63.89 & 85.19 & 73.02 \\ \cline{7-10}
		&  &  &  &     &  & NC & 80.00 & 55.17 & 65.30 \\ \cline{2-10}
		& \multirow{2}{*}{18} & \multirow{2}{*}{10} & \multirow{2}{*}{220,662} & \multirow{2}{*}{63.87} & \multirow{2}{*}{64.29} & MCI & 67.74 & 67.74 & 67.74 \\ \cline{7-10}
		&  &  &  &  &  & NC & 60.00 & 60.00 & 60.00 \\ \cline{2-10}
		& \multirow{2}{*}{19} & \multirow{2}{*}{11} & \multirow{2}{*}{252,079} & \multirow{2}{*}{76.87} & \multirow{2}{*}{76.79} & MCI & 83.87 & 76.47 & 80.00 \\ \cline{7-10}
		&  &    &  &  &  & NC & 68.00 & 77.27 & 72.34 \\ \cline{2-10}
		& \multirow{2}{*}{20} & \multirow{2}{*}{12} & \multirow{2}{*}{286,000} & \multirow{2}{*}{81.25} & \multirow{2}{*}{82.14} & MCI & 81.82 & 75.00 & 78.26 \\ \cline{7-10}
		&  &  &  &  &  & NC & 82.35 & 87.50 & 84.85 \\ \hline
		\multirow{14}{*}{\textbf{GSP block 3}} &   \multirow{2}{*}{14} & \multirow{2}{*}{6} & \multirow{2}{*}{121,048} & \multirow{2}{*}{70.75} & \multirow{2}{*}{71.43} & MCI & 66.67 & 89.66 & 76.47 \\ \cline{7-10}
		&  &  &  &  &     & NC & 82.35 & 51.85 & 63.63 \\ \cline{2-10}
		& \multirow{2}{*}{15} & \multirow{2}{*}{7} & \multirow{2}{*}{142,449} & \multirow{2}{*}{72.22} & \multirow{2}{*}{73.21} & MCI & 65.91 & 100 & 79.45 \\ \cline{7-10}
		&  &  &   &  &  & NC & 100 & 44.44 & 61.53 \\ \cline{2-10}
		& \multirow{2}{*}{16} & \multirow{2}{*}{8} & \multirow{2}{*}{166,354} & \multirow{2}{*}{67.69} & \multirow{2}{*}{67.86} & MCI & 70.00 & 70.00 & 70.00 \\ \cline{7-10}
		&  &  &  &  &  & NC & 65.38 & 65.38 & 65.38 \\ \cline{2-10}
		& \multirow{2}{*}{17} & \multirow{2}{*}{9} & \multirow{2}{*}{192,763} & \multirow{2}{*}{77.60} & \multirow{2}{*}{76.79} & MCI & 68.97 & 83.33 & 75.47 \\ \cline{7-10}
		&  &  &  &  &  & NC & 85.19 & 71.88 & 77.97 \\ \cline{2-10}
		& \multirow{2}{*}{18} & \multirow{2}{*}{10} & \multirow{2}{*}{221,676} & \multirow{2}{*}{80.14} & \multirow{2}{*}{80.36} & MCI & 78.13 & 86.21 & 81.97 \\ \cline{7-10}
		&  &     &  &  &  & NC & 83.33 & 74.07 & 78.43 \\ \cline{2-10}
		& \multirow{2}{*}{\textbf{19}} & \multirow{2}{*}{\textbf{11}} & \multirow{2}{*}{\textbf{253,093}} & \multirow{2}{*}{\textbf{88.72}} & \multirow{2}{*}{\textbf{89.29}} & MCI & \textbf{85.29} & \textbf{96.67} & \textbf{90.62} \\ \cline{7-10}
		&  &  &  &  &  & NC & \textbf{95.45} & \textbf{80.77} & \textbf{87.5} \\ \cline{2-10}
		& \multirow{2}{*}{20} & \multirow{2}{*}{12} & \multirow{2}{*}{287,014} & \multirow{2}{*}{69.74} & \multirow{2}{*}{71.43} & MCI & 66.67 & 93.33 & 77.78 \\ \cline{7-10}
		&  &  &  &  &  & NC & 85.71 & 46.15 & 60.00 \\ \hline
		\multicolumn{3}{|l|}{\multirow{2}{*}{SS-GAN \cite{SSgan} }} & \multirow{2}{*}{251,637} & \multirow{2}{*}{71.15} & \multirow{2}{*}{69.64} & MCI & 61.54 & 92.31 & 73.85 \\ \cline{7-10}
		\multicolumn{3}{|l|}{} &  &  &  & NC & 88.24 & 50.00 & 63.83 \\ \hline
		\multicolumn{3}{|l|}{\multirow{2}{*}{triple-GAN \cite{tripleGAN} }} & \multirow{2}{*}{506,386} & \multirow{2}{*}{73.44} & \multirow{2}{*}{71.43} & MCI & 61.76 & 87.50 & 72.41 \\ \cline{7-10}
		\multicolumn{3}{|l|}{} &  &  &  & NC & 86.36 & 59.38 & 70.37 \\ \hline
	\end{tabular}
	\label{table_TTrank_accuracy_MCI_NC}
\end{table*}

\begin{table*}[htbp]
	\centering
	\caption{Comparison of THS-GAN using different GSP block positions and TT-ranks for AD/MCI classification.}
	\begin{tabular}{|c|c|c|c|c|c|c|c|c|c|}
		\hline
		GSP block Position   & C\_rank & D\_rank & \#parameters & AUC(\%) & Accuracy(\%) & Class & precision(\%) & recall(\%) & f1-score(\%) \\ \hline
		\multirow{14}{*}{GSP block 1} &   \multirow{2}{*}{14} & \multirow{2}{*}{6} & \multirow{2}{*}{118,210} & \multirow{2}{*}{69.37} & \multirow{2}{*}{68.89} & AD & 62.50 & 90.91 & 74.07 \\ \cline{7-10}
		&  &  &   &  &  & MCI & 84.62 & 47.83 & 61.12 \\ \cline{2-10}
		& \multirow{2}{*}{15} & \multirow{2}{*}{7} & \multirow{2}{*}{139,611} & \multirow{2}{*}{50.00} & \multirow{2}{*}{46.94} & AD & 0 & 0 & 0 \\ \cline{7-10}
		&  &  &  &    &  & MCI & 46.94 & 100 & 63.89 \\ \cline{2-10}
		& \multirow{2}{*}{16} & \multirow{2}{*}{8} & \multirow{2}{*}{163,516} & \multirow{2}{*}{59.08} & \multirow{2}{*}{59.18} & AD & 59.09 & 54.17 & 56.52 \\ \cline{7-10}
		&  &    &  &  &  & MCI & 59.26 & 64.00 & 61.54 \\ \cline{2-10}
		& \multirow{2}{*}{17} & \multirow{2}{*}{9} & \multirow{2}{*}{189,925} & \multirow{2}{*}{63.83} & \multirow{2}{*}{64.44} & AD & 80.00 & 36.36 & 50.00 \\ \cline{7-10}
		&  &  &  &  &    & MCI & 60.00 & 91.30 & 72.41 \\ \cline{2-10}
		& \multirow{2}{*}{18} & \multirow{2}{*}{10} & \multirow{2}{*}{218,838} & \multirow{2}{*}{59.45} & \multirow{2}{*}{61.22} & AD & 70.00 & 60.43 & 64.86 \\ \cline{7-10}
		&  &  &  &  &  & MCI & 58.97 & 88.46 & 70.77 \\ \cline{2-10}
		& \multirow{2}{*}{19} & \multirow{2}{*}{11} & \multirow{2}{*}{250,255} & \multirow{2}{*}{55.18} & \multirow{2}{*}{55.10} & AD & 52.00 & 56.52 & 54.17 \\ \cline{7-10}
		&  &  &  &  &     & MCI & 58.33 & 53.85 & 56.00 \\ \cline{2-10}
		& \multirow{2}{*}{\textit{20}} & \multirow{2}{*}{\textit{12}} & \multirow{2}{*}{284,176} & \multirow{2}{*}{\textit{69.00}} & \multirow{2}{*}{\textit{71.11}} & AD & 76.92 & 50.00 & 60.61 \\ \cline{7-10}
		&  &   &  &  &  & MCI & 68.75 & 88.00 & 77.19 \\  \hline
		\multirow{14}{*}{\textbf{GSP block 2}} &   \multirow{2}{*}{14} & \multirow{2}{*}{6} & \multirow{2}{*}{120,034} & \multirow{2}{*}{63.68} & \multirow{2}{*}{67.35} & AD & 60.00 & 47.37 & 52.94 \\ \cline{7-10}
		&  &  &  &  &   & MCI & 70.59 & 80.00 & 75.00 \\ \cline{2-10}
		& \multirow{2}{*}{15} & \multirow{2}{*}{7} & \multirow{2}{*}{141,435} & \multirow{2}{*}{48.71} & \multirow{2}{*}{53.06} & AD & 38.46 & 25.00 & 30.30 \\ \cline{7-10}
		&  &  &  &     &  & MCI & 58.33 & 72.41 & 64.61 \\ \cline{2-10}
		& \multirow{2}{*}{16} & \multirow{2}{*}{8} & \multirow{2}{*}{165,340} & \multirow{2}{*}{70.65} & \multirow{2}{*}{69.39} & AD & 86.67 & 50.00 & 63.42 \\ \cline{7-10}
		&  &  &  &     &  & MCI & 61.76 & 91.30 & 73.68 \\ \cline{2-10}
		& \multirow{2}{*}{\textbf{17}} & \multirow{2}{*}{\textbf{9}} & \multirow{2}{*}{\textbf{191,749}} & \multirow{2}{*}{\textbf{85.35}} & \multirow{2}{*}{\textbf{85.71}} & AD & \textbf{85.71} & \textbf{88.89} & \textbf{87.27} \\ \cline{7-10}
		&  &  &     &  &  & MCI & \textbf{85.71} & \textbf{81.82} & \textbf{83.72} \\ \cline{2-10}
		& \multirow{2}{*}{18} & \multirow{2}{*}{10} & \multirow{2}{*}{220,662} & \multirow{2}{*}{61.45} & \multirow{2}{*}{61.22} & AD & 56.00 & 63.64 & 59.58 \\ \cline{7-10}
		&  &  &  &  &    & MCI & 66.67 & 59.26 & 62.75 \\ \cline{2-10}
		& \multirow{2}{*}{19} & \multirow{2}{*}{11} & \multirow{2}{*}{252,079} & \multirow{2}{*}{57.74} & \multirow{2}{*}{63.27} & AD & 80.00 & 19.05 & 30.77 \\ \cline{7-10}
		&  &  &  &  &     & MCI & 61.36 & 96.43 & 75.00 \\ \cline{2-10}
		& \multirow{2}{*}{20} & \multirow{2}{*}{12} & \multirow{2}{*}{286,000} & \multirow{2}{*}{45.26} & \multirow{2}{*}{48.98} & AD & 33.33 & 25.00 & 28.57 \\ \cline{7-10}
		&  &  &  &  &    & MCI & 55.88 & 65.52 & 60.32 \\  \hline
		\multirow{14}{*}{GSP block 3} &   \multirow{2}{*}{14} & \multirow{2}{*}{6} & \multirow{2}{*}{121,048} & \multirow{2}{*}{63.44} & \multirow{2}{*}{71.43} & AD & 70.73 & 93.55 & 80.56 \\ \cline{7-10}
		&  &  &  &  &    & MCI & 75.00 & 33.33 & 46.15 \\ \cline{2-10}
		& \multirow{2}{*}{\textit{15}} & \multirow{2}{*}{\textit{7}} & \multirow{2}{*}{142,449} & \multirow{2}{*}{\textit{74.00}} & \multirow{2}{*}{\textit{73.47}} & AD & 80.95 & 65.38 & 72.34 \\ \cline{7-10}
		&  &    &  &  &  & MCI & 67.86 & 82.61 & 74.51 \\ \cline{2-10}
		& \multirow{2}{*}{16} & \multirow{2}{*}{8} & \multirow{2}{*}{166,354} & \multirow{2}{*}{72.07} & \multirow{2}{*}{71.43} & AD & 80.00 & 61.54 & 69.57 \\ \cline{7-10}
		&  &  &  &  &   & MCI & 65.52 & 82.61 & 73.08 \\ \cline{2-10}
		& \multirow{2}{*}{17} & \multirow{2}{*}{9} & \multirow{2}{*}{192,763} & \multirow{2}{*}{59.47} & \multirow{2}{*}{55.10} & AD & 75.00 & 40.00 & 52.17 \\ \cline{7-10}
		&  &  &  &  &     & MCI & 45.45 & 78.95 & 57.69 \\ \cline{2-10}
		& \multirow{2}{*}{18} & \multirow{2}{*}{10} & \multirow{2}{*}{221,676} & \multirow{2}{*}{49.56} & \multirow{2}{*}{57.14} & AD & 37.50 & 15.79 & 22.22 \\ \cline{7-10}
		&  &  &    &  &  & MCI & 60.98 & 83.33 & 70.42 \\ \cline{2-10}
		& \multirow{2}{*}{19} & \multirow{2}{*}{11} & \multirow{2}{*}{253,093} & \multirow{2}{*}{69.05} & \multirow{2}{*}{71.43} & AD & 71.00 & 86.00 & 77.78 \\ \cline{7-10}
		&  &  &  &  &    & MCI & 73.00 & 52.00 & 60.74 \\ \cline{2-10}
		& \multirow{2}{*}{20} & \multirow{2}{*}{12} & \multirow{2}{*}{287,014} & \multirow{2}{*}{70.37} & \multirow{2}{*}{67.35} & AD & 100 & 41.00 & 58.16 \\ \cline{7-10}
		&  &  &  &    &  & MCI & 58.00 & 100 & 73.42 \\    \hline
		\multicolumn{3}{|l|}{\multirow{2}{*}{SS-GAN \cite{SSgan} }} & \multirow{2}{*}{251,637} & \multirow{2}{*}{50.00} & \multirow{2}{*}{48.98} & AD & 48.98 & 100 & 65.75 \\ \cline{7-10}
		\multicolumn{3}{|l|}{} &  &  &  & MCI & 0 & 0 & 0 \\ \hline
		\multicolumn{3}{|l|}{\multirow{2}{*}{triple-GAN \cite{tripleGAN} }} & \multirow{2}{*}{506,386} & \multirow{2}{*}{72.14} & \multirow{2}{*}{73.47} & AD & 76.47 & 59.09 & 66.67 \\ \cline{7-10}
		\multicolumn{3}{|l|}{} &  &  &  & MCI & 71.88 & 85.19 & 77.97 \\ \hline
	\end{tabular}
	\label{table_TTrank_accuracy_AD_MCI}
\end{table*}

To investigate the effect of TT-rank and different GSP block position on classification performance, this section provides a comparative evaluation of the proposed THS-GAN with respect to a range of TT-rank values and different GSP block positions for each evaluation group.
The TT-core number is fixed as 3. As far as we know, there have been no published studies that adopt tensor-train decomposition in GAN for semi-supervised classification. Thus the most suitable TT-rank remains to be explored. Nonetheless, we conducted a variety of preliminary experiments, and have empirically chosen TT-ranks according to the performances in our validation sets. More specifically, we consider the effect of TT-ranks on classification performance  when  $\mathrm{C}_{-} \mathrm{rank}=\{14,15,16,17,18,19,20\}$ and $\mathrm{D}_{-} \mathrm{rank}=\{6,7,8,9,10,11,12\}$. Note that   $\mathrm{C}_{-} \mathrm{rank}$ and $\mathrm{D}_{-}\mathrm{rank}$ represent TT-rank of classifier and discriminator respectively.
SS-GAN \cite{SSgan} and triple-GAN \cite{tripleGAN} are used as two baseline models for comparison purpose.
With respect to SS-GAN, the discriminator has 3 output units corresponding to $\text { [CLASS-1, CLASS-} 2, \text {FAKE}]$. CLASS-1 and CLASS-2 correspond to one of classes AD, MCI, NC respectively according to the evaluation group. In this case, discriminator can also act as classifier.   For the fair comparison, the two baselines have the same structure and hyperparameter settings as our model but without tensor-train decomposition and high-order module GSP block.

From Table \ref{table_TTrank_accuracy_AD_NC}, it can be observed that the best AUC can be achieved using  $\mathrm{C}_{-} \mathrm{rank}=20$ and $\mathrm{D}_{-} \mathrm{rank}=12$ no matter the GSP block is at the position of either GSP block 1, GSP block 2 or GSP block 3 in the context
of AD/NC classification. The best AUC of 95.92\% is obtained when GSP block 2 is inserted after Dense-BC block 2. On the other hand, in the context of MCI/NC classification,  Table \ref{table_TTrank_accuracy_MCI_NC} shows that the optimal TT-rank  is not consistent with AD/NC classification when GSP block is positioned at different locations.  With respect to GSP block 1, a good AUC of 85.71\% is obtained when $\mathrm{C}_{-} \mathrm{rank}=18$ and $\mathrm{D}_{-} \mathrm{rank}=10$. Similarly regarding GSP block 2, a good AUC of 88.32\% is obtained when   $\mathrm{C}_{-} \mathrm{rank}=15$ and $\mathrm{D}_{-} \mathrm{rank}=7$. In the same manner, with respect to GSP block 3, a good AUC of 88.72\% is obtained when   $\mathrm{C}_{-} \mathrm{rank}=19$ and $\mathrm{D}_{-} \mathrm{rank}=11$. The best AUC of 88.72\% is obtained when GSP block 3 is utilized. In the context of AD/MCI classification,  Table \ref{table_TTrank_accuracy_AD_MCI} also indicates the same trend that the optimal TT-rank is different when GSP block is positioned at different locations. With respect to GSP block 1, a good AUC of 69.37\% is obtained when  $\mathrm{C}_{-} \mathrm{rank}=14$ and $\mathrm{D}_{-} \mathrm{rank}=6$. Similarly regarding GSP block 2, a good AUC of 85.35\% is obtained when  $\mathrm{C}_{-} \mathrm{rank}=17$ and $\mathrm{D}_{-} \mathrm{rank}=9$. In the same manner, with respect to GSP block 3, a good AUC of 74\% is obtained when  $\mathrm{C}_{-} \mathrm{rank}=15$ and $\mathrm{D}_{-} \mathrm{rank}=7$. The best AUC of 85.35\% is obtained when GSP block 2 is utilized.

From Table \ref{table_TTrank_accuracy_AD_NC} to Table \ref{table_TTrank_accuracy_AD_MCI}, the following overall observations can be made.
(1)  THS-GAN with optimal hyperparameter settings can achieve the best classification performance in terms of AUC and accuracy compared with triple-GAN and SS-GAN. The triple-GAN performs better than SS-GAN, which confirms that the triple-GAN can alleviate the competing problem of SS-GAN
that the discriminator has two incompatible convergence points. (2) Compared with the triple-GAN, THS-GAN
can obtain AUC gains of 9.09\% (95.92\%-86.83\%) for AD/NC classification, 15.28\% (88.72\%-73.44\% ) for MCI/NC classification,  and 13.21\% (85.35\%-72.14\%) for AD/MCI classification, improving the performance by a large margin. This indicates that the performance of the proposed model is significantly improved by introducing tensor-train decomposition and high-order pooling. Furthermore, THS-GAN used far fewer parameters, compared with the triple-GAN which  used 506,386 parameters. The compression rates are  506,386/286,000 =1.77 for AD/NC classification, 506,386/253,093 =2  for MCI/NC classification, and 506,386/191,749 =2.64 for AD/MCI classification respectively.
(3) According to our results,
the best classification results are obtained by utilizing either GSP block 2 or GSP block 3, but not GSP block 1.
This observation indicates that exploiting the second-order statistics in the later layers can improve the predictive power significantly.
The conjectured reason for this is that the features extracted in the earlier layers are simple and common, but in the later layers representative features will be abstracted,  and by inserting the high-order pooling module GSP block in the later layers,  more discriminative features can be enhanced and redundant features will be suppressed; thus the predictive performance is improved.   Although inserting GSP block
at the later layers will increase the number of parameters, the best trade-off between accuracy and number of parameters should be
chosen at GSP block 2. GSP block 2 arrangement leads to
the best accuracy with the optimal TT-ranks.  (4) TT-rank has a significant effect on testing accuracy, and the optimal  value of TT-rank depends on network architecture and data. It is difficult to specify an optimal value for TT-rank in advance. Again, this observation is consistent with \cite{novikov15tensornet} that finding optimal TT-rank remains a challenge.
According to the experimental results, the optimal value of TT-rank lies in the range
$[14,20]$ for classifier and $[6,12]$ for discriminator. It is not time-consuming to find it in practical applications.
Under optimal TT-ranks, THS-GAN can achieve better performance than triple-GAN and our model uses fewer parameters, which indicates that TT-decomposition can utilize parameters more efficiently, and is less likely to converge to local minima. Note that the optimal hyperparameter settings  for each evaluation group will be utilized in the rest of the experiments.

\subsection{The effect of the amount of labeled data}

\begin{figure}
	\includegraphics[width=\linewidth]{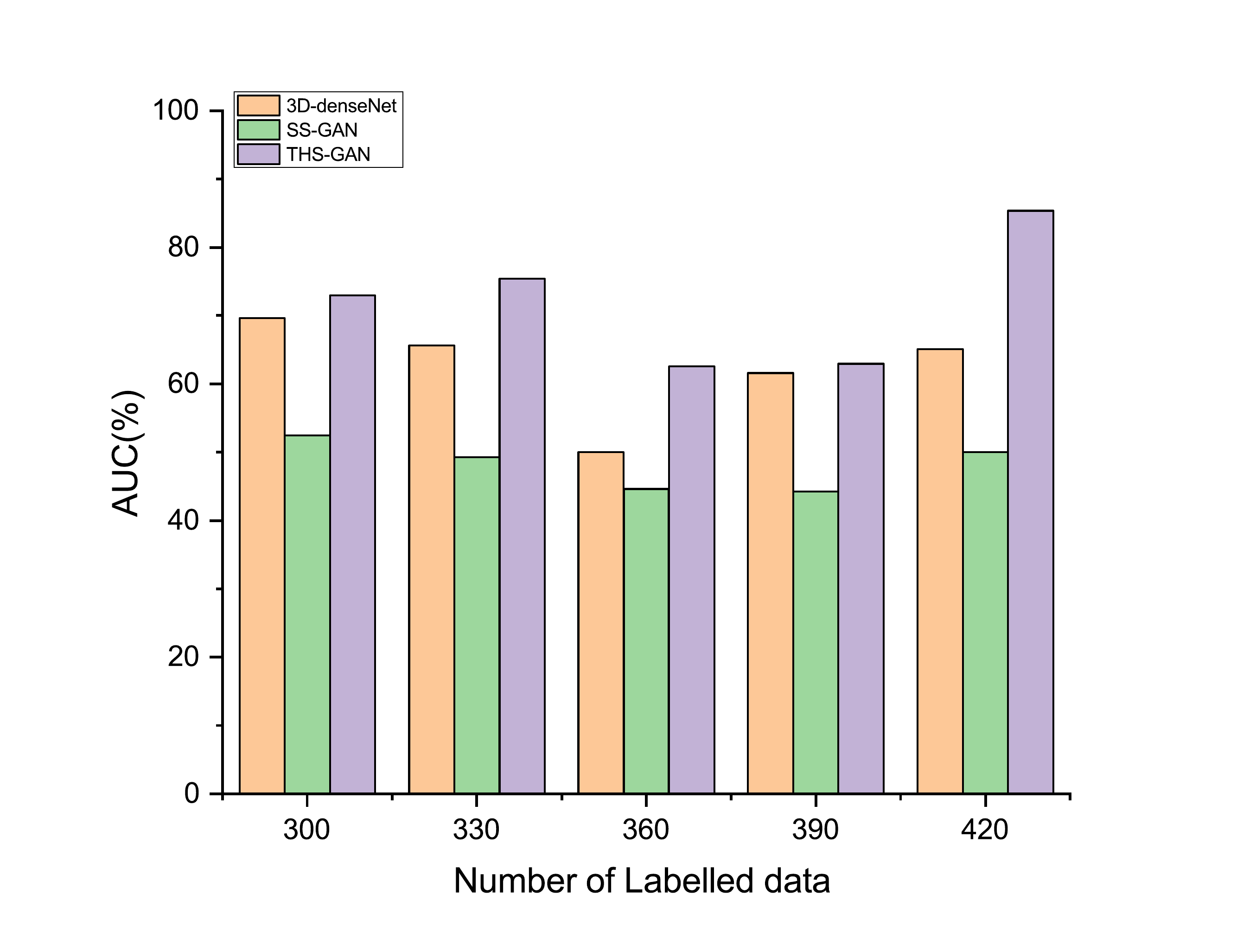}
	\caption{Comparison of different number of labeled data for AD/MCI classification.}
	\label{fig_AD_MCI_lessData}
\end{figure}

\begin{figure}
	\includegraphics[width=\linewidth]{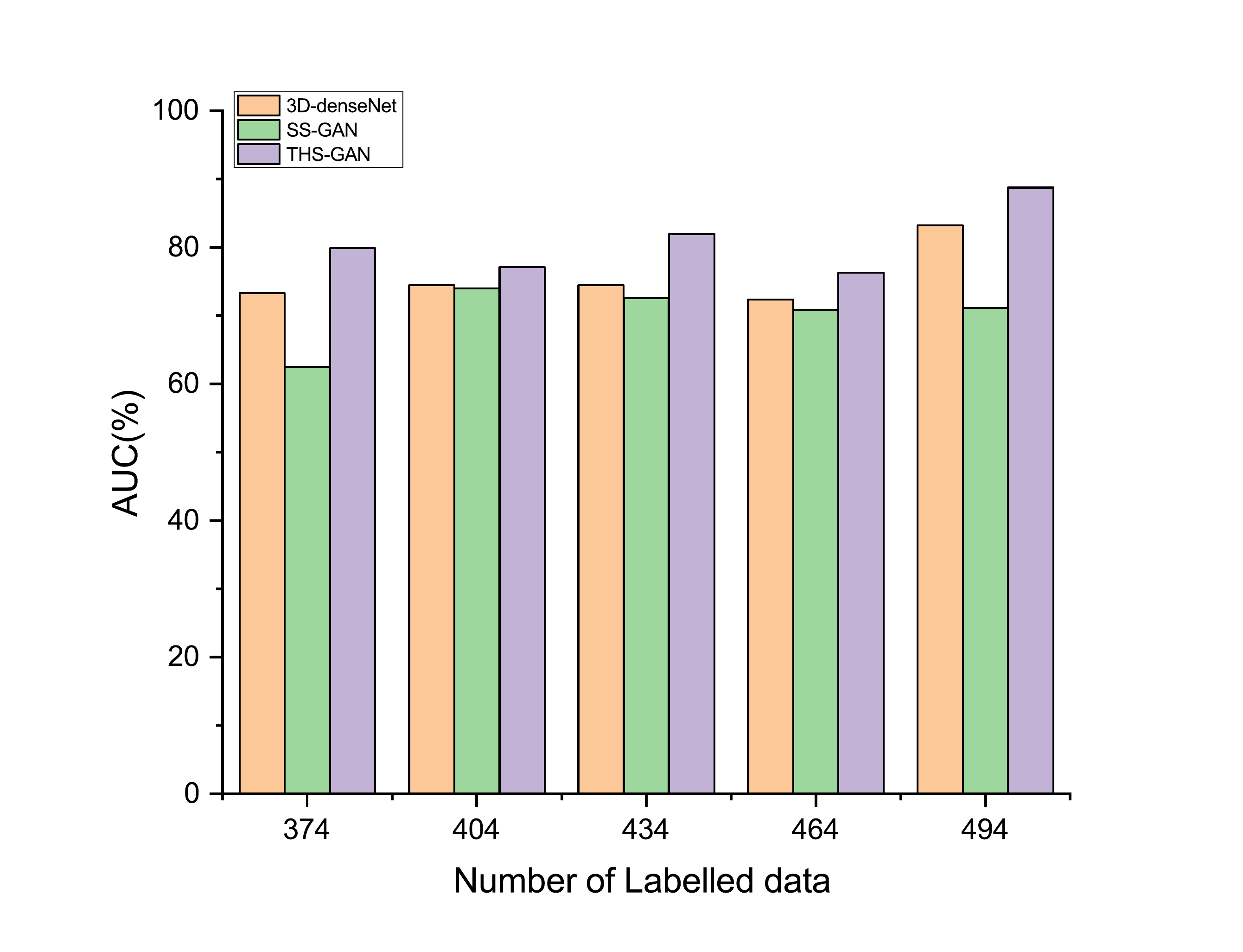}
	\caption{Comparison of different number of labeled data for MCI/NC classification.}
	\label{fig_MCI_NC_lessData}
\end{figure}

\begin{figure}
	\includegraphics[width=\linewidth]{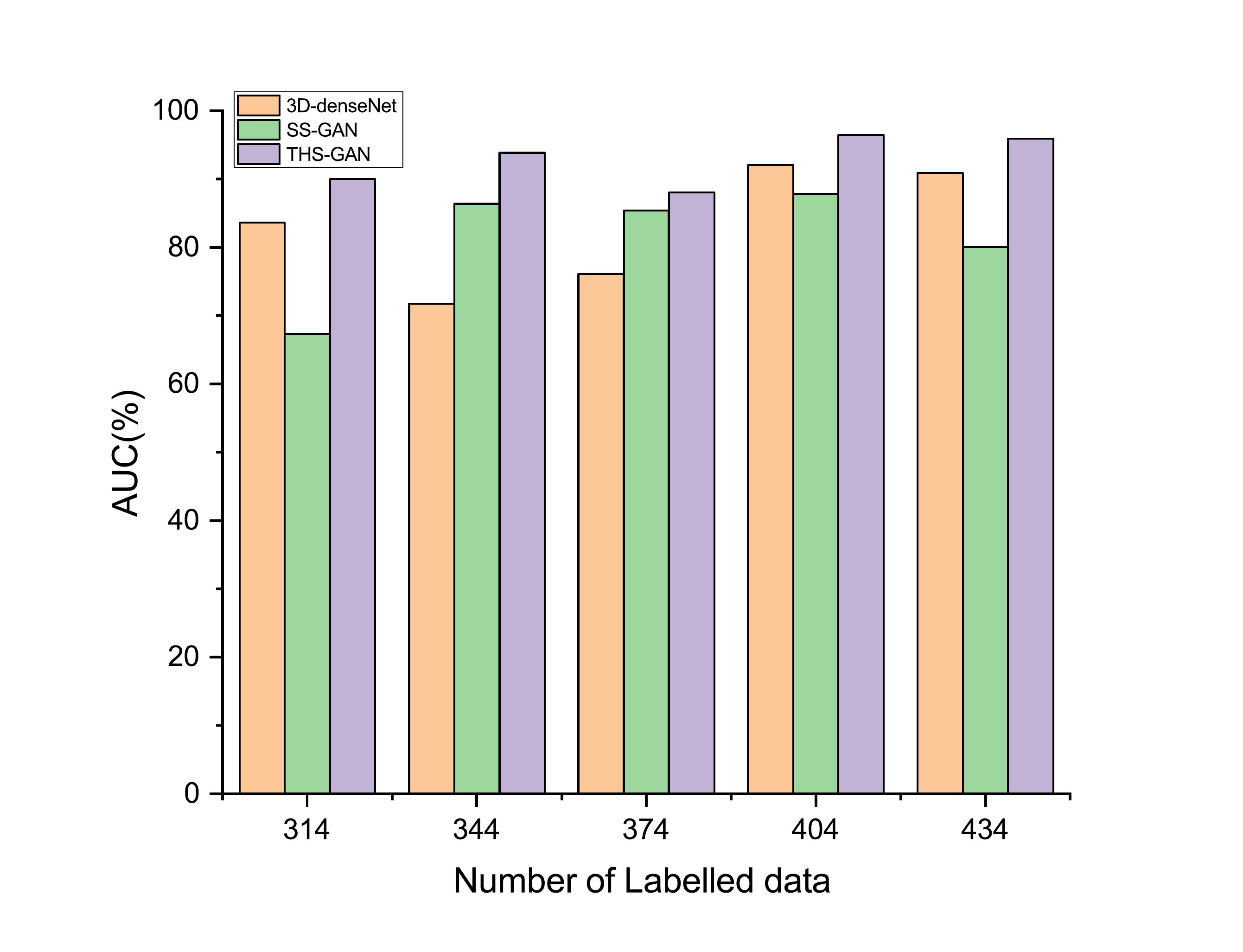}
	\caption{Comparison of different number of labeled data for AD/NC classification.}
	\label{fig_AD_NC_lessData}
\end{figure}

In this subsection, we investigate the effect of using the different number of labeled data for semi-supervised classification.  For our proposed THS-GAN, the architecture and hyperparameters are fixed as the optimal settings found in Section \ref{experimentTTrank}.
The 3D-DenseNet architecture is the same as the classifier of THS-GAN
but without tensor-train decomposition and GSP block. Similarly, the structure of SS-GAN is also the same as THS-GAN
but without tensor-train decomposition and GSP block.   It can be seen from Fig.\ref{fig_AD_MCI_lessData} that as the number of labeled data increased, our THS-GAN outperforms SS-GAN by a large margin and performs better than 3D-DenseNet when there are less labeled data for AD/MCI classification.  Fig. \ref{fig_MCI_NC_lessData} shows that as the number of labeled data increased, our THS-GAN always outperforms both 3D-DenseNet and SS-GAN for MCI/NC classification.
The same trend can be found in Fig. \ref{fig_AD_NC_lessData}. We can also observe that the THS-GAN requires fewer labeled samples to achieve comparable results. In Fig. \ref{fig_AD_MCI_lessData}, when the number of labeled data is small such as 300, our THS-GAN
can still achieve better performance than SS-GAN and 3D-DenseNet which use more labeled data such as 330, 360, 390 and 420 respectively in the context of AD/MCI classification. Similar trends can also be found for MCI/NC and AD/NC  in Fig. \ref{fig_MCI_NC_lessData} and Fig. \ref{fig_AD_NC_lessData} respectively.  This improvement is given by the real MRI images without labels and the synthetic MRI images produced by the generator.

\subsection{The effect of number of parameters}

\begin{figure}
	\includegraphics[width=\linewidth]{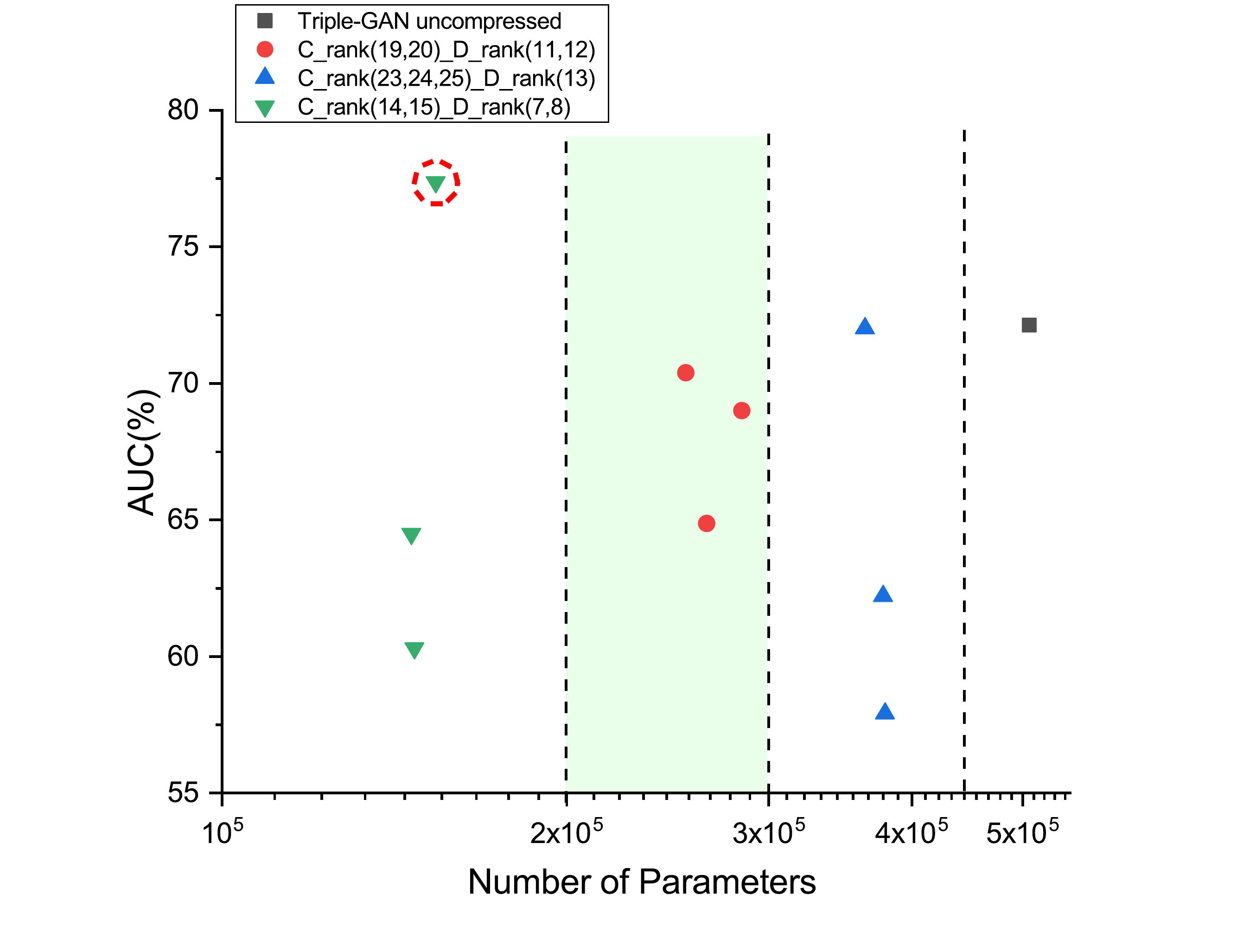}
	\caption{Comparison of different number of parameters for AD/MCI classification.}
	\label{fig_AD_MCI_parameters}
\end{figure}

In this subsection, we investigate the properties of THS-GAN and compare with triple-GAN uncompressed in the context of AD/MCI classification. In order to compare the performance for the same range of parameters, varies TT-ranks are utilized with respect to THS-GAN.
The result in Fig. \ref{fig_AD_MCI_parameters} illustrates that THS-GAN
can obtain the best AUC with optimal TT-ranks when the number of parameters is compressed in the range [$10^{5}$, $2 \times 10^{5}$](in red dashed circle).  Furthermore, THS-GAN can achieve comparable AUC when TT-ranks are set to large numbers,
and the number of parameters is in the range of [$2 \times 10^{5}$, $3 \times 10^{5}$] or  [$3 \times 10^{5}$, $4 \times 10^{5}$].  Overall speaking, THS-GAN can achieve much better performance when TT-rank is set to be not large, and the number of the parameter is compressed between $10^{5}$ and  $2 \times 10^{5}$ so that the predictive performance will be boosted.

\subsection{The convergence comparison}
\begin{figure}[ht] 
	\includegraphics[width=\linewidth]{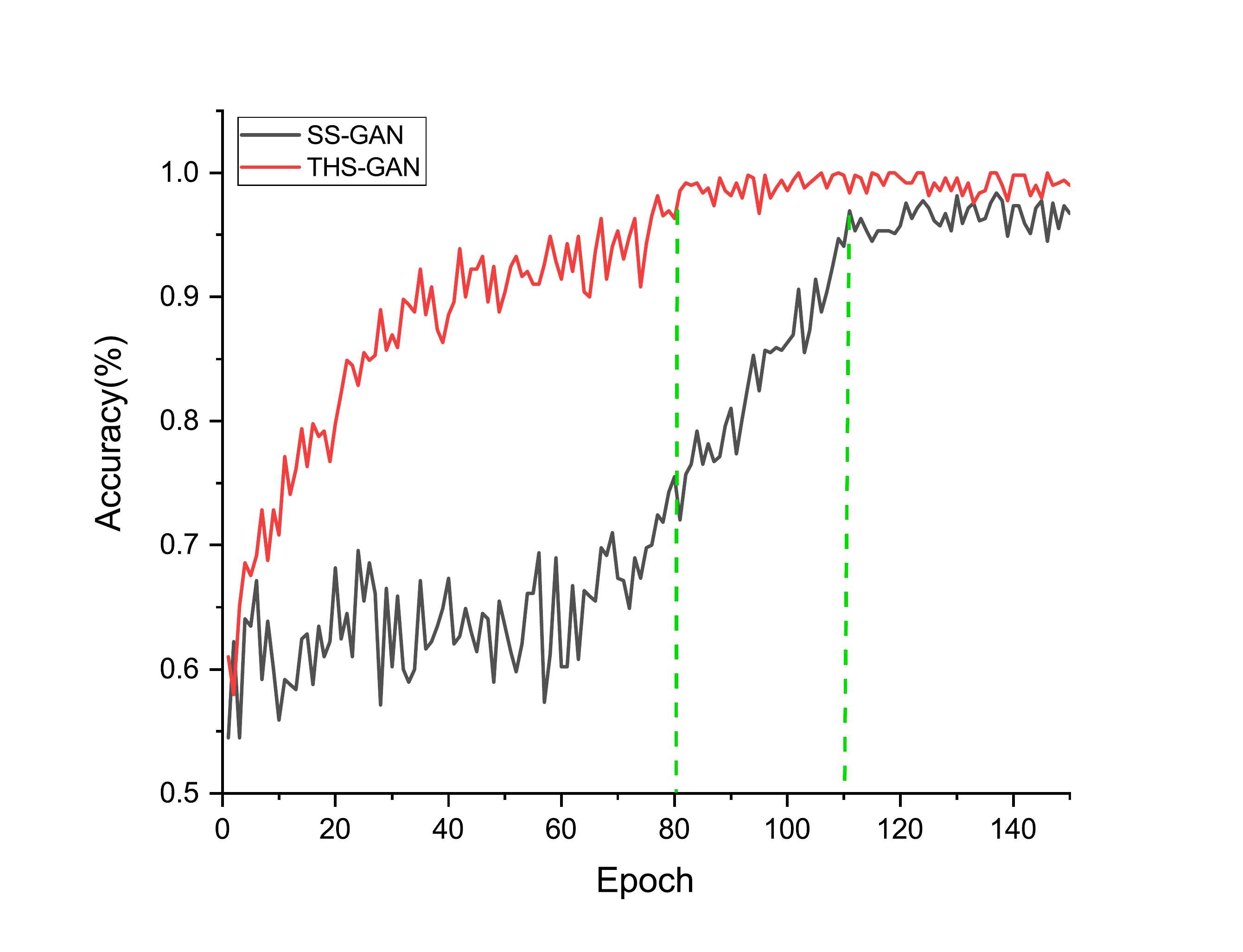}
	\caption{Convergence curves for MCI/NC  classification}
	\label{fig_MCI_NC}
\end{figure}
\begin{figure}[ht] 
	\includegraphics[width=\linewidth]{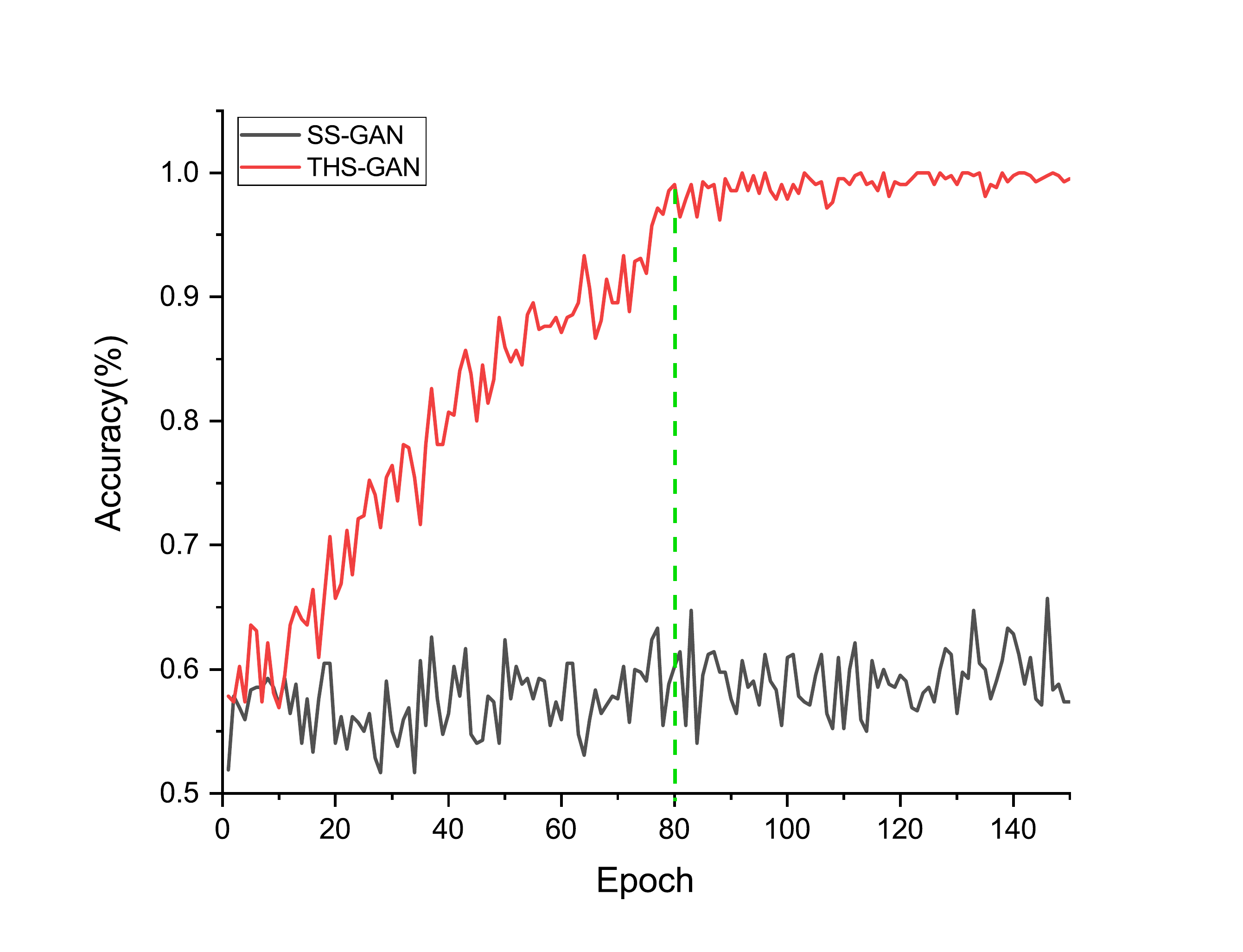}
	\caption{Convergence curves for AD/MCI classification}
	\label{fig_AD_MCI}
\end{figure}
Fig. \ref{fig_MCI_NC} and Fig. \ref{fig_AD_MCI} show the convergence curves of our THS-GAN and SS-GAN
for evaluation group MCI/NC and AD/MCI  respectively.  Our model converges faster than conventional SS-GAN. In the case of AD/MCI classification, SSgan can not converge since the differences between AD and MCI are so subtle that the T1-MRI images of AD, MCI, and fake are hard to be distinguished by the discriminator. These results are also consistent with Table \ref{table_TTrank_accuracy_MCI_NC} and Table  \ref{table_TTrank_accuracy_AD_MCI} in Section \ref{experimentTTrank}. More specifically, for MCI/NC classification, AUC of THS-GAN (88.72\%) is much higher than SS-GAN
(71.15\%) since THS-GAN converges faster than SS-GAN. For AD/MCI classification, AUC of SS-GAN
is only 50\%, and SS-GAN can not converge during the training process. On the other hand, our THS-GAN converges faster and AUC of 85.35\% can be achieved.

\subsection{The visualization of generated images}
\begin{table*}[bp]
	\centering
	\begin{tabular}{c p{1.8cm}<{\centering} c p{1.7cm}<{\centering} p{1.8cm}<{\centering} c p{1.7cm}<{\centering} p{1.8cm}<{\centering} c p{1.7cm}<{\centering} }
		&\multicolumn{3}{c}{30 Epoches}  & \multicolumn{3}{c}{60 Epoches} & \multicolumn{3}{c}{90 Epoches}    \\  
		&Coronal  &           Sagittal & Axial & Coronal  & Sagittal & Axial & Coronal  & Sagittal & Axial    \\
		SS-GAN  &\multicolumn{3}{l}{\includegraphics[width=0.3\linewidth]{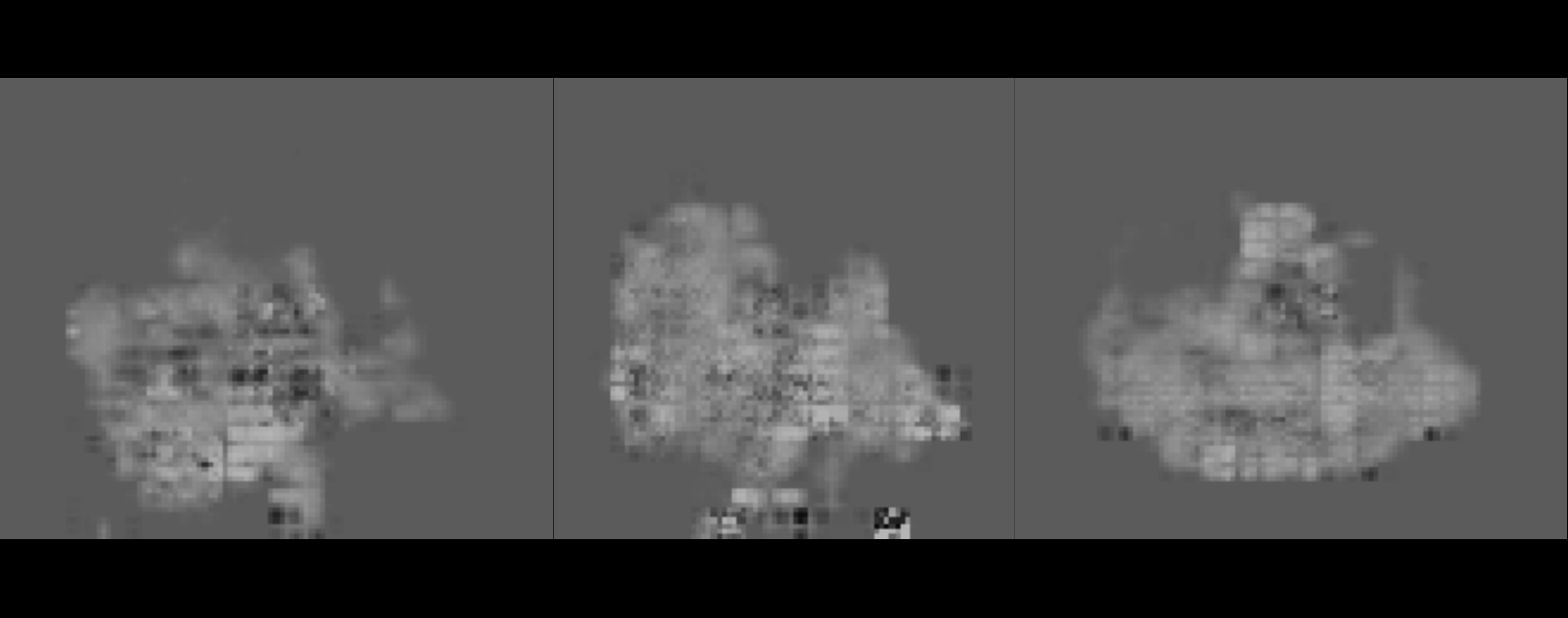}}  &
		\multicolumn{3}{l}{\includegraphics[width=0.3\linewidth]{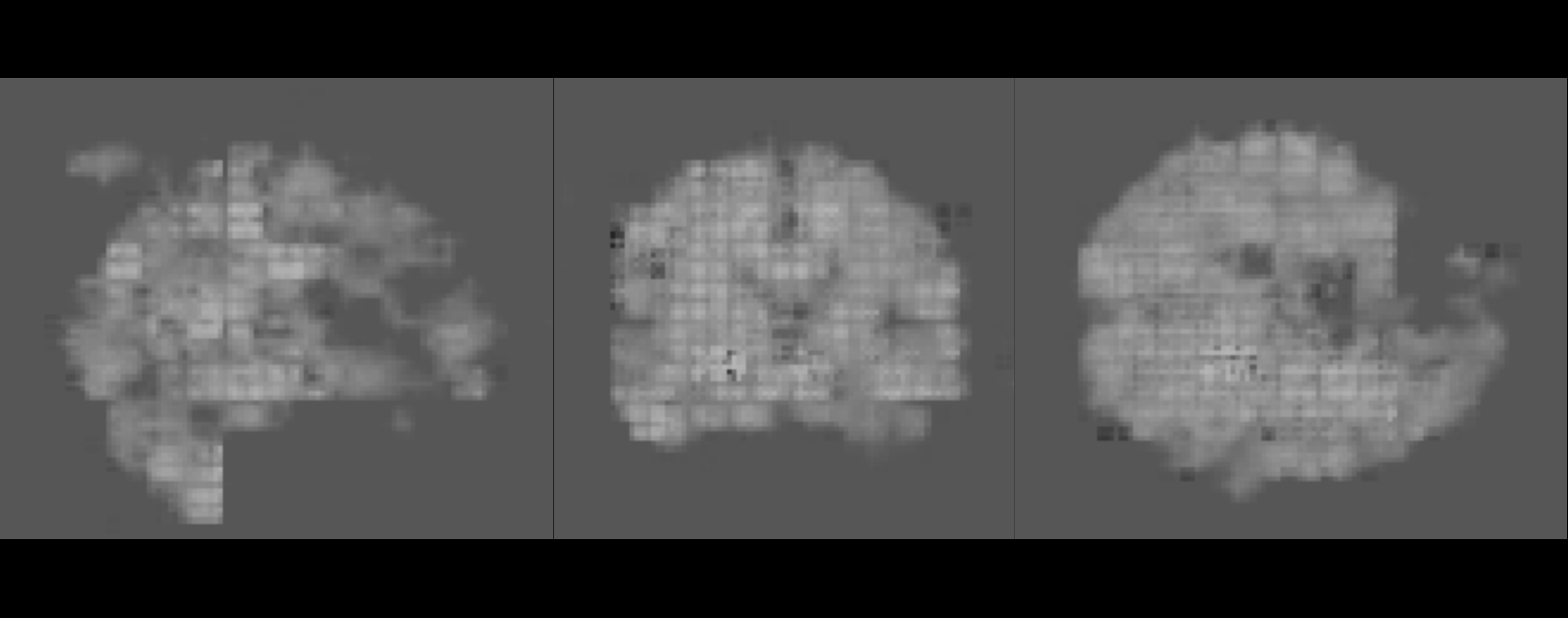}} &
		\multicolumn{3}{l}{\includegraphics[width=0.3\linewidth]{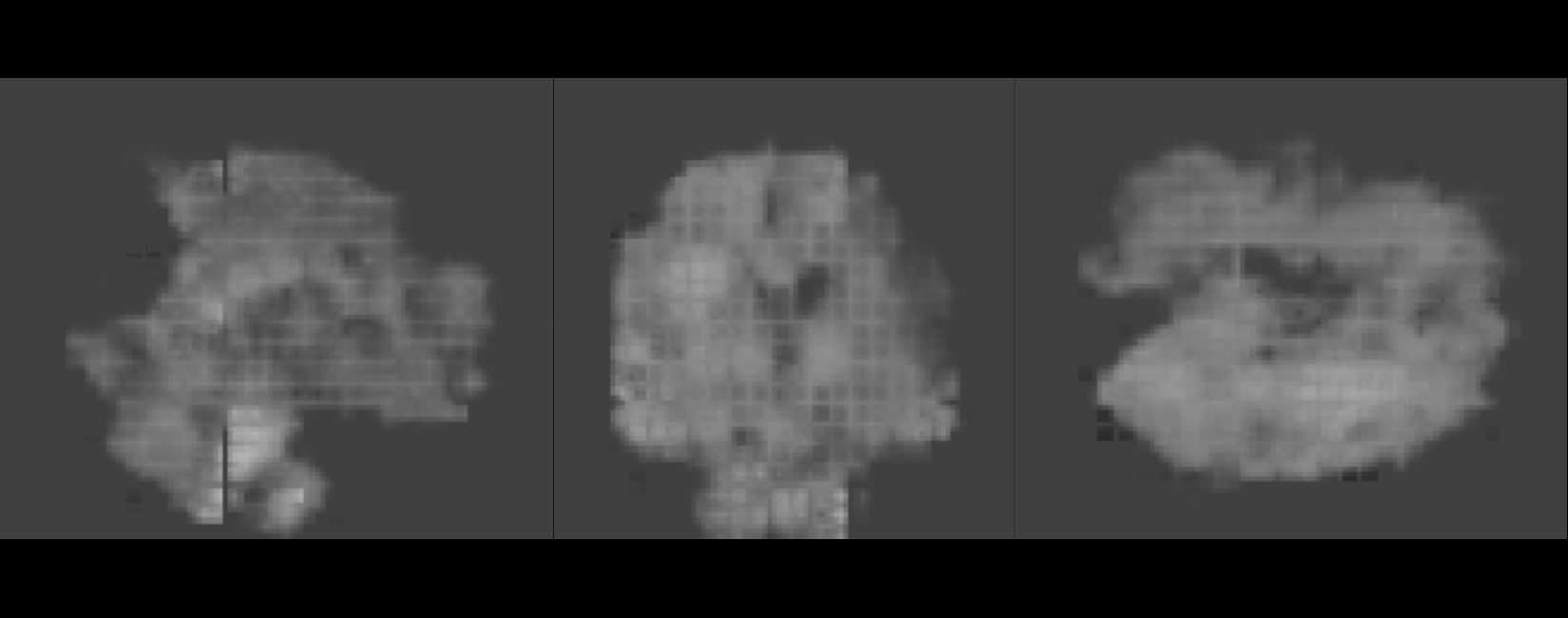}}  \\  
		\hline
		
		GAN  &\multicolumn{3}{l}{\includegraphics[width=0.3\linewidth]{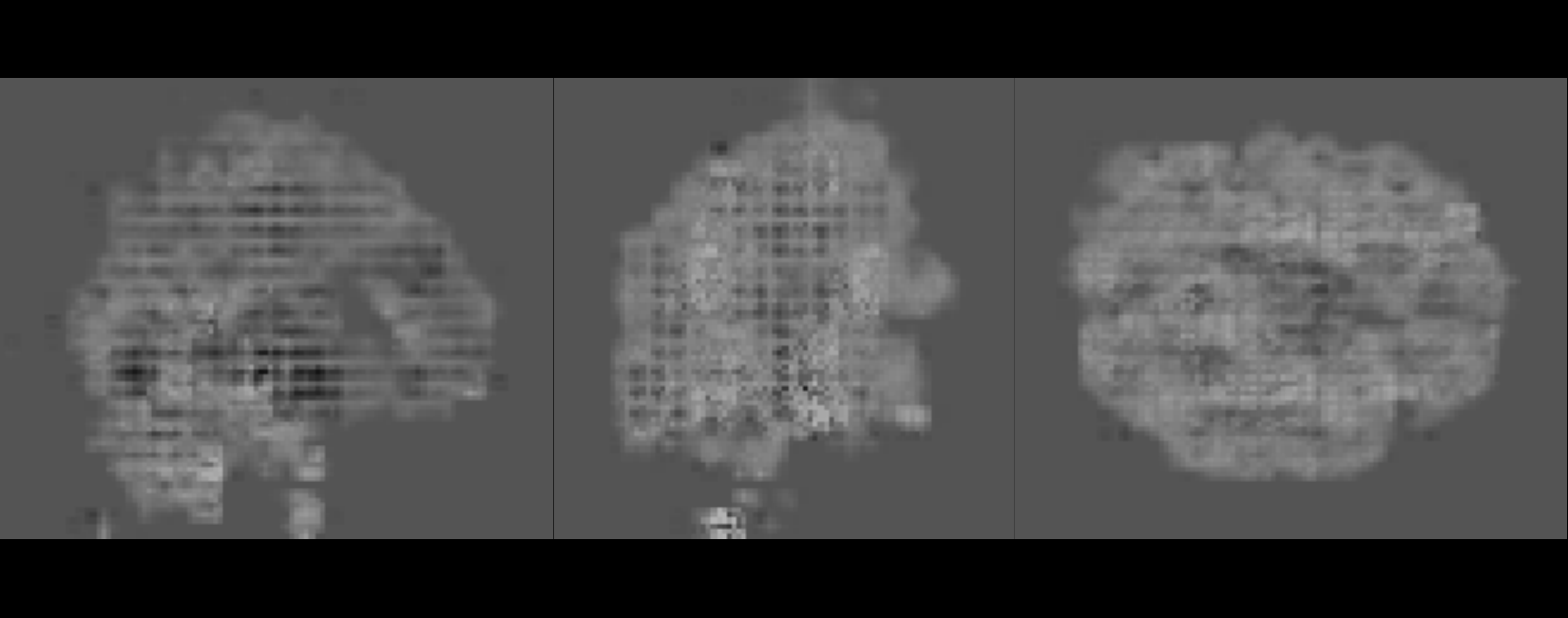}}  &
		\multicolumn{3}{l}{\includegraphics[width=0.3\linewidth]{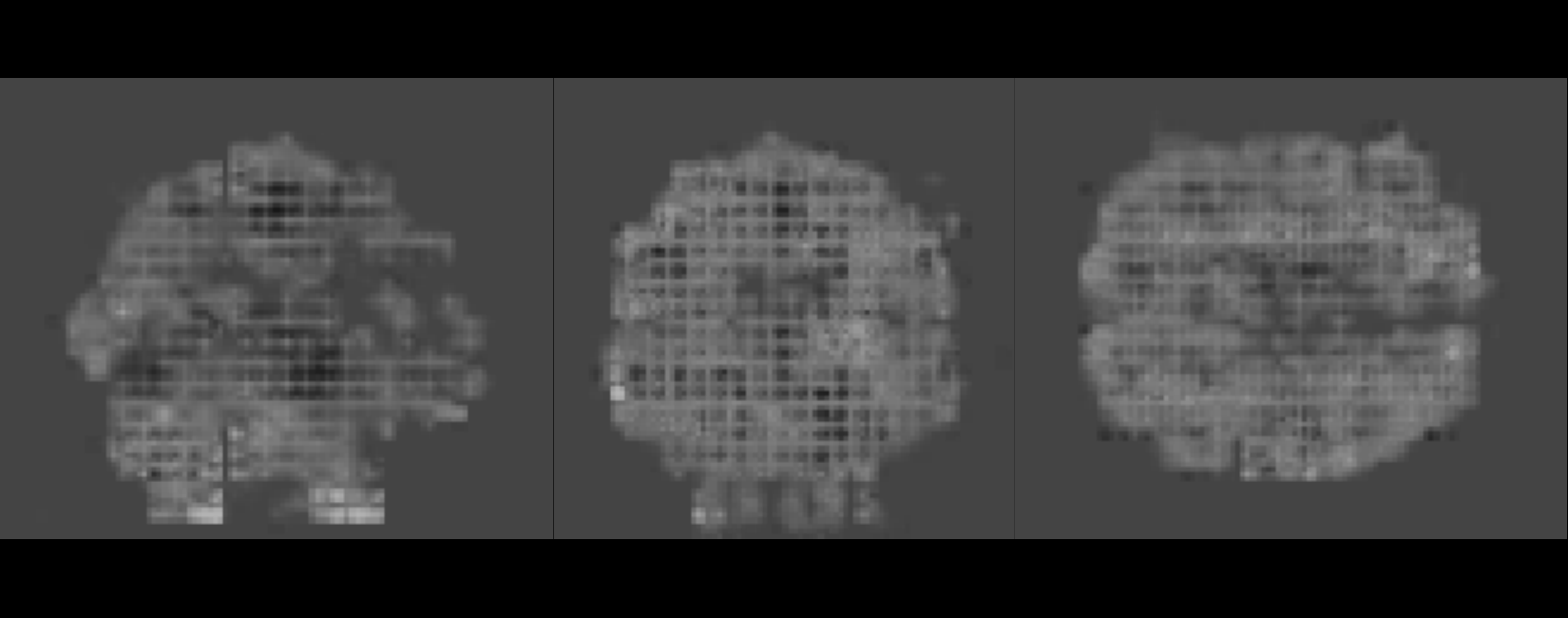}} &
		\multicolumn{3}{l}{\includegraphics[width=0.3\linewidth]{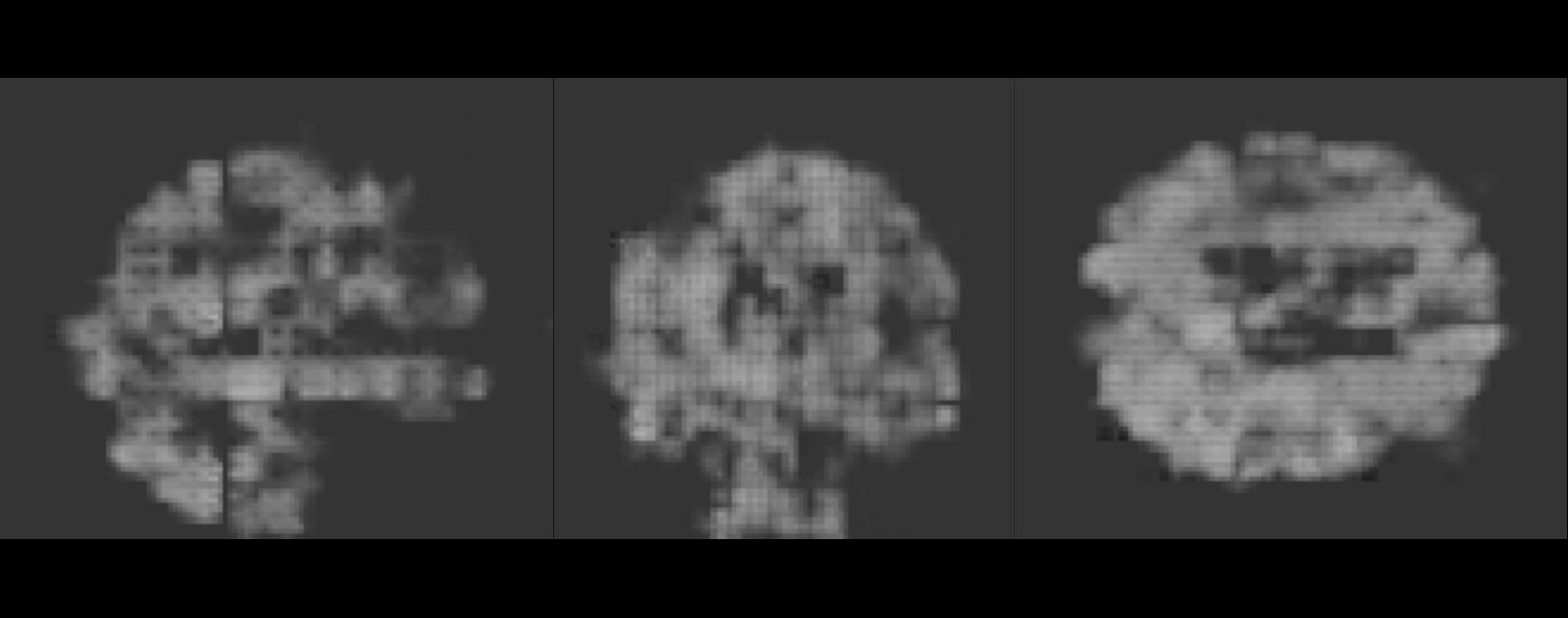}}   \\   
		\hline
		THS-GAN  &\multicolumn{3}{l}{\includegraphics[width=0.3\linewidth]{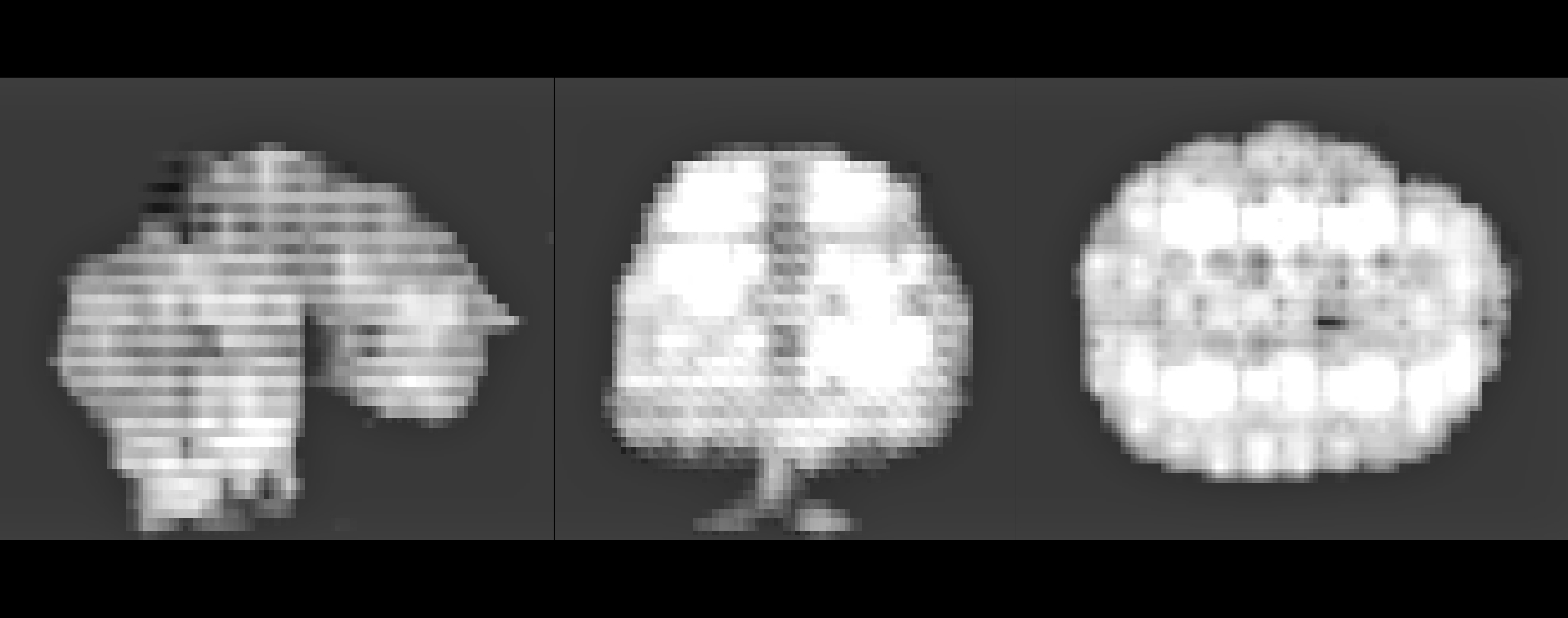}}  &
		\multicolumn{3}{l}{\includegraphics[width=0.3\linewidth]{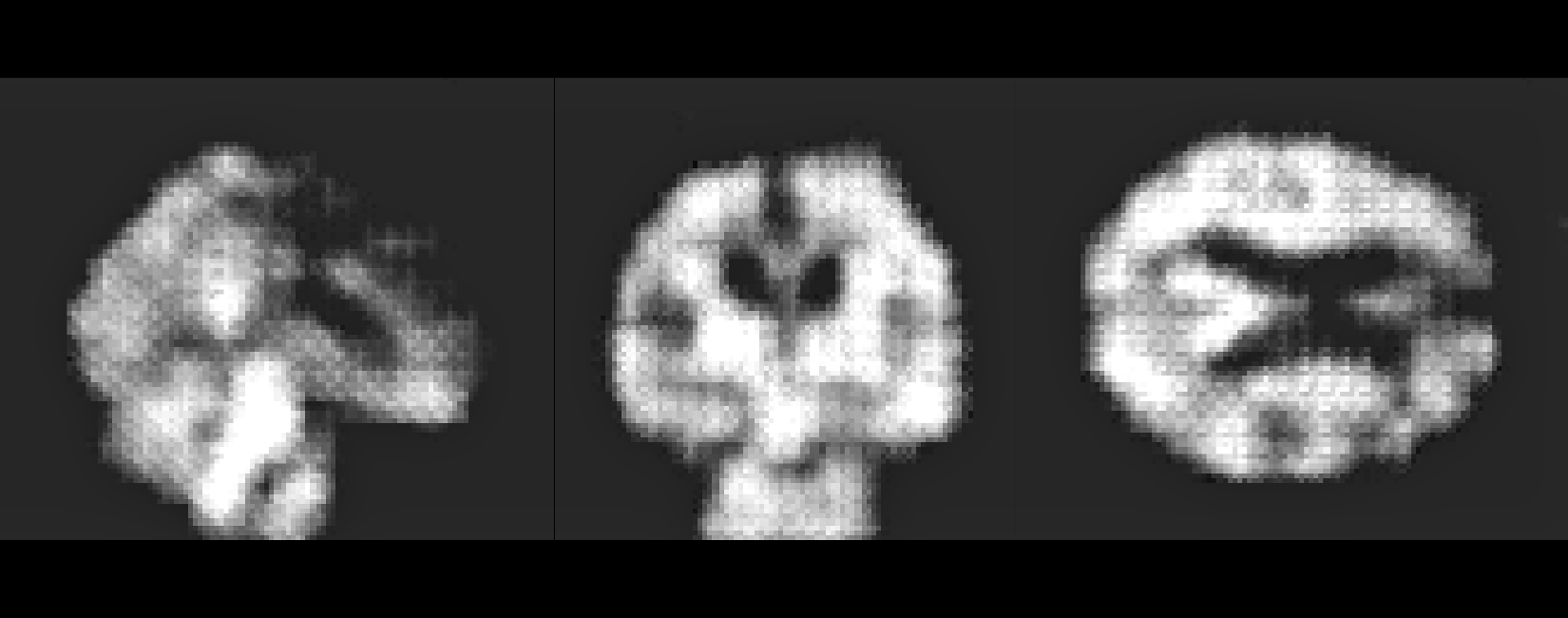}}   &
		\multicolumn{3}{l}{\includegraphics[width=0.3\linewidth]{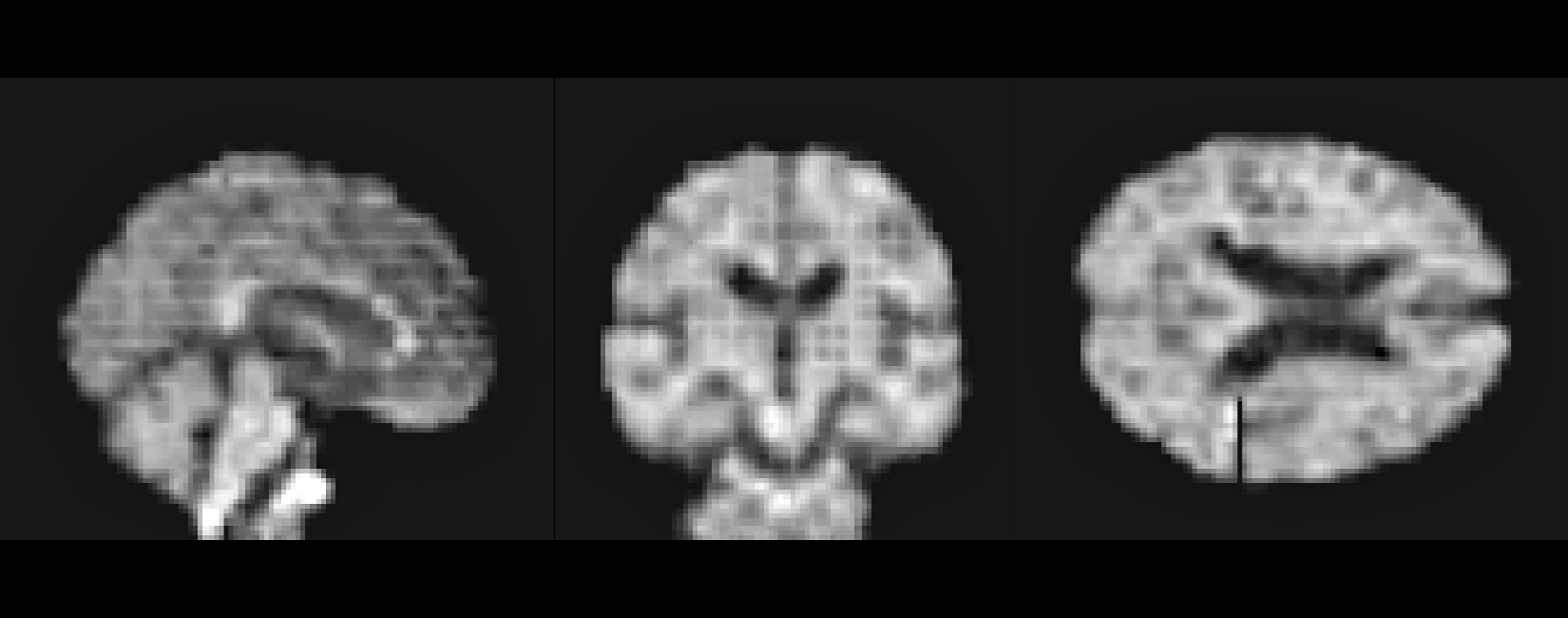}} \\   
		\hline
		Real  &\multicolumn{3}{l}{\includegraphics[width=0.3\linewidth]{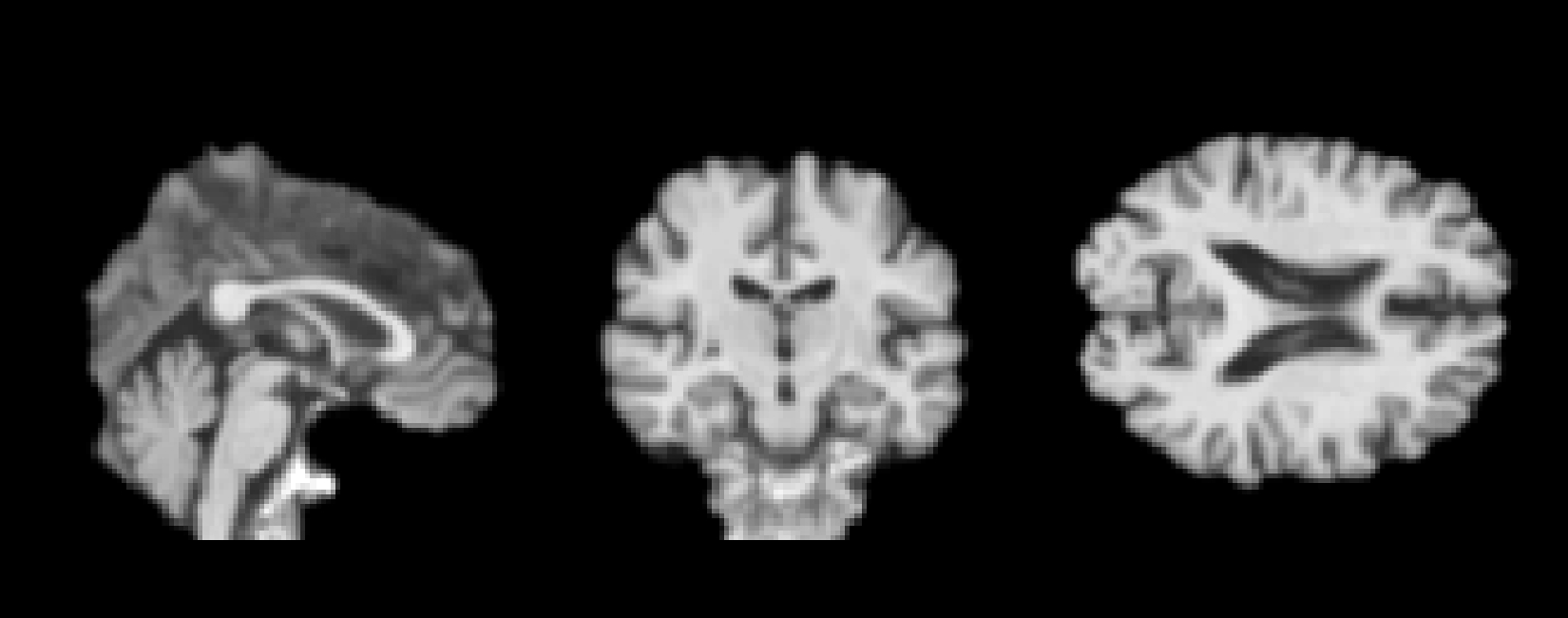}}	  &
		\multicolumn{3}{l}{\includegraphics[width=0.3\linewidth]{ac.pdf}} &
		\multicolumn{3}{l}{\includegraphics[width=0.3\linewidth]{ac.pdf}}   \\
		\multicolumn{10}{c}{   \qquad   \qquad        Fig.11:  Comparison of brain MRI slices generated by SS-GAN, GAN and THS-GAN.}
	\end{tabular}
\end{table*}

In this subsection, we visualize the center-cut slices of the generated 3D T1-MRI images from random latent vectors during the training process as shown in Fig. 11.  In the beginning, generated samples are blurry, and the detailed features of the brain disappear. In the latter stage, compared with SS-GAN and GAN, the generated samples from our proposed THS-GAN can reflect more detailed attributes of the brain (e.g., sulci, gyri).

\subsection{The comparison with existing methods}
\begin{table*}
	\centering
	\caption{Comparison with existing methods}	
	\begin{tabular}{|c|c|c|c|c|c|c|c|c|c|}
		\hline
		\multirow{2}{*}{Model} &  \multicolumn{3}{c|}{MCI vs NC(\%)} & \multicolumn{3}{c|}{AD vs MCI(\%)} & \multicolumn{3}{c|}{AD vs NC(\%)} \\ \cline{2-10}
		& ACC & Recall & AUC & ACC & Recall & AUC & ACC & Recall & AUC \\ \hline
		Plocharski et al. \cite{featureSVM} &   84.40 & 82.30 & 84.00 & 81.50 & 81.70 & 83.00 & 92.30 & 91.30 & 98.00 \\ \hline
		Peng et al. \cite{MKL} &   71.60 & 83.90 & - & 65.40 & 41.20 & - & 88.40 & 84.10 & - \\ \hline
		Xu et al. \cite{regressionSVM}  & 70.89 & 61.39 & 79.02 & - & - & - & 90.40 & 92.36 & 95.36 \\ \hline
		Neffati et al. \cite{DKPCA_SVM}  & - & - & - & - & - & - & 91.11 & 85.00 & - \\ \hline
		Li et al. \cite{Patch3DdenseNetRNN}   & 75.00 & 81.90 & 75.80 & - & - & - & 89.10 & 84.60 & 91.00 \\ \hline
		Cui et al. \cite{CNN_RNN}  & - & - & - & - & - & - & 91.33 & 86.87 & 93.22 \\ \hline
		Ren et al. \cite{2Dcnn}   & 88.50 & 82.16 & 82.00 & 85.32 & 78.79 & 80.00 & 93.75 & 94.23 & 93.00 \\ \hline
		Liu et al. \cite{compare8}  & 77.84 & 76.81 & 82.72 & - & - & - & 84.97 & 82.65 & 90.63 \\ \hline
		Cheng et al. \cite{compare3}  & - & - & - & - & - & - & 87.15 & 86.36 & 92.26 \\ \hline
		THS-GAN   &\textbf{ 89.29} & \textbf{96.67} & \textbf{88.72} & \textbf{85.71} & \textbf{88.89} & \textbf{85.35} & \textbf{95.92} & \textbf{95.83} &  95.92  \\ \hline
	\end{tabular}
	\label{table_otherMethods}
\end{table*}
Several machine learning methods have been tried for the discrimination of subjects using structural MRI images. Table \ref{table_otherMethods}  presents the reported performance of some related studies. Although a direct comparison of these studies is difficult, as each study uses different datasets and preprocessing protocols, the table indicates comparison results for the classification of T1-MRI images. Table \ref{table_otherMethods} demonstrates that our proposed method performs better than the previous methods.   Compared with Plocharski et al. \cite{featureSVM}, our proposed method requires less image-preprocessing steps for feature extraction. No segmentation and rigid registration are required in our method.  In particular,  Li et al. \cite{Patch3DdenseNetRNN} constructed denseNets on the decomposed image patches of the internal and external hippocampus to learn the intensity and shape features. Then Recurrent neural network (RNN) is cascaded to combine the features from the left and right hippocampus, then the high-level features are abstracted for disease classification.   Cheng et al. \cite{compare3} proposed a method based on a combination of multiple 3D-CNNs to classify AD and NC subjects. It is built on different local image patches to transform the local brain image into more compact high-level features.  Our method outperforms those methods as well. It demonstrates the benefit of tensor-train decomposition and the high-order pooling module leveraged in our THS-GAN. Our method achieves superior classification performance, indicating its potential capability of assessing MCI and AD.

\section{Conclusions}\label{Conclusions}

In this paper, we developed a novel THS-GAN for assessing  MCI and AD.  The three-player cooperative game based framework is tensorized so that the proposed model can benefit from the structural information of the brain. By introducing high-order pooling in our model,  more significant features can be extracted by making full use of the second-order statistics of the holistic MRI images. Thus the capability of our model is enhanced.   To the best of our knowledge, the proposed THS-GAN model is the first work to consider tensor-train decomposition in GAN and leverage GAN for
semi-supervised classification on MRI images for AD diagnosis. The experimental results demonstrate that the proposed THS-GAN model can obtain promising results. The visualization of the generated images during the training process also shows that our model can generate plausible MRI images. In future work, we will investigate the generated MRI images for data augmentation.

%

\section*{Acknowledgment}

This work is supported by National Natural Science Foundations of China under Grant No. 61872351, Shenzhen Key Basic Research Project under Grant No.JCYJ20180507182506416.

\ifCLASSOPTIONcaptionsoff
  \newpage
\fi

\bibliographystyle{IEEEtran}
\bibliography{reference}

\end{document}